\documentclass{article}
    \PassOptionsToPackage{numbers, compress}{natbib}

    \usepackage[final]{neurips_2021}

\usepackage[utf8]{inputenc} 
\usepackage[T1]{fontenc}    
\usepackage{url}            
\usepackage{amsfonts}       
\usepackage{nicefrac}       
\usepackage{microtype}      

\usepackage{tikz}
\usetikzlibrary{positioning, shapes.geometric}
\usepackage{bbding}
\usepackage{times}
\usepackage[toc,page]{appendix}
\usepackage{latexsym}
\usepackage{natbib}
\usepackage{graphicx}
\usepackage{subfigure}
\usepackage{amsmath}
\usepackage{amsthm}
\usepackage{amsfonts,amssymb}
\usepackage{booktabs}
\usepackage{multicol}
\usepackage{multirow}
\usepackage{diagbox}
\usepackage{ulem}
\usepackage{bm}
\usepackage{pifont}
\DeclareMathOperator*{\argmax}{\arg\!\max}
\usepackage{cleveref}
\crefname{section}{§}{§§}
\Crefname{section}{§}{§§}
\usepackage{dashrule}
\newtheorem{innercustomgeneric}{\customgenericname}
\providecommand{\customgenericname}{}
\newcommand{\newcustomtheorem}[2]{%
  \newenvironment{#1}[1]
  {%
   \renewcommand\customgenericname{#2}%
   \renewcommand\theinnercustomgeneric{##1}%
   \innercustomgeneric
  }
  {\endinnercustomgeneric}
}
\usepackage{caption}

\newcustomtheorem{customthm}{Theorem}
\newcustomtheorem{customcoro}{Corollary}

\usepackage[linesnumbered,boxed,ruled,commentsnumbered]{algorithm2e}
\usepackage{lipsum}

\newcommand{\tabincell}[2]{\begin{tabular}{@{}#1@{}}#2\end{tabular}}

\title{Global-aware Beam Search for Neural Abstractive Summarization}

\author{Ye Ma$^{1,4}$ \qquad   Zixun Lan$^{2,4}$ \qquad   Lu Zong$^{1,4}$\thanks{corresponding author}\qquad   Kaizhu Huang$^3$\\
\\
$^1$\ Department of Financial and Actuarial Mathematics, School of Science\\
$^2$\ Department of Applied Mathematics, School of Science\\
$^3$\ Department of Intelligent Science, School of Advanced Technology\\
$^4$\ Laboratory for Intelligent Computing and Financial Technology\\
Xi’an Jiaotong-Liverpool University,
SIP, 215123 Suzhou, China\\
\\
\{ye.ma, lu.zong, kaizhu.huang\}@xjtlu.edu.cn, zixun.lan19@student.xjtlu.edu.cn}

\begin{document}
\maketitle
\begin{abstract}

This study develops a calibrated beam-based algorithm with awareness of the global attention distribution for neural abstractive summarization, aiming to improve the local optimality problem of the original beam search in a rigorous way. Specifically, a novel global protocol is proposed based on the attention distribution to stipulate how a global optimal hypothesis should attend to the source. A global scoring mechanism is then developed to regulate beam search to generate summaries in a near-global optimal fashion. This novel design enjoys a distinctive property, i.e., the global attention distribution could be predicted before inference, enabling step-wise improvements on the beam search through the global scoring mechanism. Extensive experiments on nine datasets show that the global (attention)-aware inference significantly improves state-of-the-art summarization models even using empirical hyper-parameters. The algorithm is also proven robust as it remains to generate meaningful texts with corrupted attention distributions. The codes and a comprehensive set of examples are available.\footnote{\url{https://github.com/yema2018/global_aware}}
\end{abstract}

\section{Introduction}\label{sec:int}

As the barriers exist from the auto-regressive design of neural probabilistic text generators to predicting the global optimum directly~\cite{seq2seq14}, the heuristic algorithm beam search that factorizes global optimization to multiple local optimizations, has been popularly used for text decoding~\cite{meister2020if}. In the original beam search setting, the global optimum is a hypothesis $\mathbf{g}$ of the highest probability among all possible sentences, and consists of words in vocabulary $\mathcal{V}$. Given the global optimum at step $t$ denoted as $\mathbf{g}_{\leq t}$, the local optimums $\mathbf{l}_{\leq t}$ refer to $\mathcal{K}$ candidate sequences with the highest probabilities at each step. While it is necessary to compromise on the beam size $\mathcal{K} \ll \mathcal{V}$ to ensure text quality \cite{cohen2019empirical,ott2018analyzing,meister2020if} and search efficiency, beam search suffers from a major limitation  due to its local property. Concretely, assuming that the global optimal hypothesis is within the $\mathcal{K}$ local optimal hypotheses of the highest probabilities, i.e. $p(\mathbf{g}_{\leq t}) \geq p(\mathbf{l}_{\leq t})$, for all $t$ until the termination $T$, it operates solely with the local information available at each step. In practice, such assumption may however fail in the case that the probability of the global optimum at step $\tau<T$ is less than those of the local optimums, i.e. $p(\mathbf{g}_{\leq \tau}) < p(\mathbf{l}_{\leq \tau})$, but is adjusted to a higher level in the later steps, $p(\mathbf{g}_{> \tau}|\mathbf{g}_{\leq \tau}) > p(\mathbf{l}_{> \tau}|\mathbf{l}_{\leq \tau})\ \mathrm{and} \  p(\mathbf{g}_{\leq \tau})p(\mathbf{g}_{> \tau}|\mathbf{g}_{\leq \tau}) > p(\mathbf{l}_{\leq \tau})p(\mathbf{l}_{> \tau}|\mathbf{l}_{\leq \tau})$. This often leads beam search to get stuck in the local optimum from step $\tau$ onward in generating texts.

\begin{figure}
\centering
\includegraphics[scale=0.6]{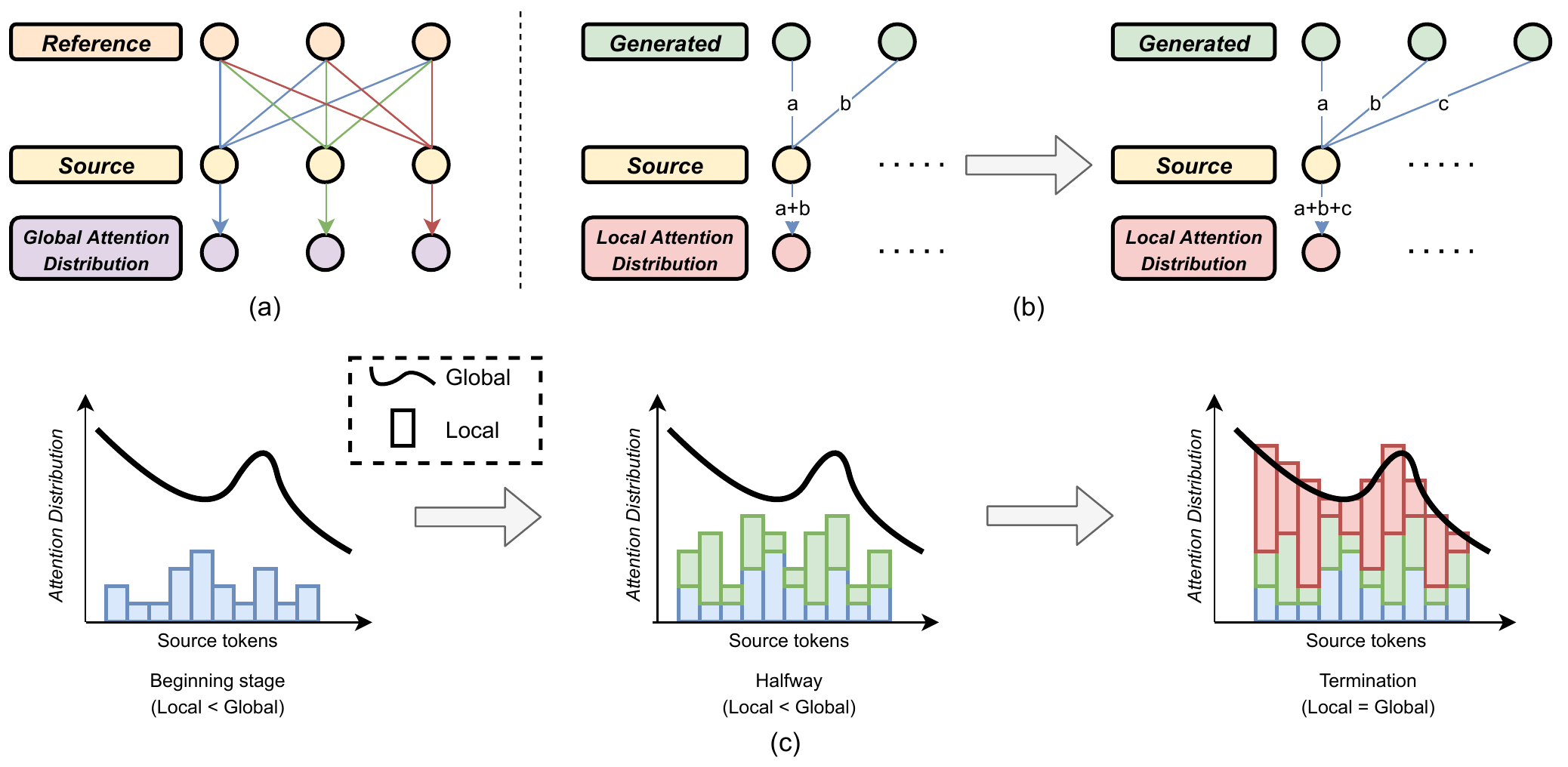}
\caption{(a) Attention distribution is composed of the summation of cross attention on the same-colored lines, distinguished from that of different-colored lines which always equals $1$ due to softmax. (b) Local attention gradually increases as the decoding proceeds. (c) Desired situation: growing local attention has been lower than global attention during decoding and exactly reaches it at the end.}
\label{fig:att-aware}
\end{figure}

To cope with this limitation, this study proposes a calibrated beam-based algorithm with global awareness at all searching steps. Generally, our novel algorithm is implemented in two phases. Before the beam search (Phase I), the \textit{global attention distribution} is predicted in order to be included as a protocol to calibrate beam search at each step, encouraging the generated hypotheses to attend to the source in a more near-global optimal way. Specifically, the global attention distribution describes how  all reference tokens should assign the attention to each source token (illustrated in Figure \ref{fig:att-aware}.a), which could be predicted from the source by training an attention-prediction model. The training is fairly straightforward and resembles a sequence tagging task~\cite{huang2015bidirectional}, except that the predicted attention distribution from the source is a regression result. There are several advantages of using the attention distribution as the global protocol. 1) Attention distributions are sensitive to the decoder input, suggesting that any input to the decoder leads to a unique attention distribution with fixed model parameters; 2) attention distributions are accessible for almost all text generation tasks thanks to the recent advances in attention models~\cite{radford2019language, devlin2018bert, vaswani2017attention};  3) relying on the source only, the global attention distribution can be predicted before beam search, thus offering a rigorous mechanism to calibrate a global-aware beam search.

During beam search (Phase II), we develop a novel \textit{global scoring mechanism} composed of attention scores and length rewards to guide beam search based on the predicted global attention distribution. As one main theoretical result, we show that the attention score can be considered as the probability that generated texts attend to sources in line with the predicted global attention distribution. Specifically, the generated tokens in each step update the local attention distribution to source tokens dynamically, where the attention values grow monotonically as the generation proceeds (see Figure \ref{fig:att-aware}.b). Since the desired situation is that the local distribution reaches exactly the global distribution at the terminal step, we regulate the inference by discouraging local attention from exceeding their corresponding predicted global attention at all steps.

With regards to the core target to investigate the possible paradigm that improves beam search with global awareness during decoding, contributions of this study are summarized as follows:
\begin{itemize}
    \item We argue that the limitation of beam search roots from its defect in finding the global optimal hypothesis. We improve the algorithm by proposing a global protocol to regulate beam search step-by-step. This paper is the first to predict and deploy the global attention distribution to calibrate the inference in a rigorous way, thus  returning a hypothesis that attends to source tokens in a more near-global optimal manner. In contrast, previous works~\cite{Wu2016Google,gehrmann-etal-2018-bottom, li-etal-2018-improving-neural,See2017, li2018simple} try to use attention distributions to improve beam search, but ignore that the global attention distribution is predictable.
    
    \item A novel global scoring mechanism is designed to evaluate the generated sequences at each step based on the desired situation described in Figure \ref{fig:att-aware}.c.
    As theoretically justified, its major component can be elegantly integrated into beam search in the form of a probability so that merely $\mathcal{O}(\mathcal{K})$ of the time complexity is increased in each step (see Section \cref{sec:proposed} for more details). 
    
    \item The proposed algorithm with global awareness manifests a robust and plug-and-play property in enhancing beam search for neural abstractive summarization. Without requiring any model or parameter modification, the global-aware inference shows excellent performance in generating meaningful texts, even if the attention distribution  is corrupted or not of its own. Further, it is identified that summaries generated by global-aware inference are both higher-quality and  different from beam search hypotheses (see {\it Global-aware} in Table \ref{tab:SAMPLE}). More interestingly, we find that the generation style of a dataset could be transferred by the designated global attention distribution. For instance, summaries of higher abstractness for CNN/DM could be generated by only replacing its global attention distribution with a highly abstractive distribution during inference, as presented in {\it Global-aware$\dagger$} of Table \ref{tab:SAMPLE}.
    
    \item  On the empirical side, we show that the proposed global-aware inference can stably and significantly boost two state-of-the-art summarization models BART \cite{lewis-etal-2020-bart} and PEGASUS \cite{Zhang2020PEGASUSPW} to produce higher quality summaries on nine datasets,  even if only the empirical hyper-parameters are used.

\end{itemize}

\begin{table}[htbp]\scriptsize
    \centering
     \caption{Use BART \cite{lewis-etal-2020-bart} fine-tuned in CNN/DM to generate summaries. {\it Global-aware} uses the attention distribution learned from CNN/DM, while {\it Global-aware$\dagger$} takes the attention distribution learned from XSUM.}
    \begin{tabular}{p{43pt}|p{330pt}}
    \toprule
      {\it Beam search}&President Obama says climate change is a public health issue that affects all of us . Obama: "No challenge poses more of a public threat than climate change" Obama: "Millions of people would lose their health insurance" if Affordable Care Act is not upheld . Obama: "I am not anticipating the Supreme Court would make such a bad decision"\\
      \midrule
      {\it Global-aware} & President Obama says climate change is a public health issue that affects all of us . He says the average American can do their part to reduce their own carbon footprint . Obama did not appear particularly concerned about the Supreme Court challenge to the Affordable Care Act .\\
       \midrule
      {\it Global-aware$\dagger$} & President Barack Obama says climate change is a public health issue . He says the average American can do their part to reduce their carbon footprint .\\
        \bottomrule
    \end{tabular}
    
    \label{tab:SAMPLE}
\end{table}

\section{Preliminary}
The proposed decoding strategy is applied in BART~\cite{lewis-etal-2020-bart} and PEGASUS~\cite{Zhang2020PEGASUSPW} to perform summarization. BART is a pre-trained seq-to-seq model whose structure essentially follows a vanilla Transformer encoder-decoder~\cite{vaswani2017attention}. PEGASUS has a similar structure but is pre-trained on a larger dataset differently. Notably, the fine-tuned parameters of both models are downloaded from {\it HuggingFace Models}\footnote{https://huggingface.co/models} and are fixed in all subsequent operations, where ``fine-tuned" means the pre-trained model has been fine-tuned on a specific dataset.

\subsection{Beam Search}
 Since the decoder of the seq-to-seq model is an auto-regressive model, the probability of a target sequence $\hm{y}=(y_0, \cdots,y_t,\cdots, y_T)$ can be factorized to the probabilities conditional on the source $\hm{x} = (x_1,\cdots,x_i,\cdots,x_n)$ and $\hm{y}_{<t}=(y_0, \cdots, y_{t-1})$, i.e., 
\begin{equation}
    p(\hm{y}| \hm{x}) = \prod_{t=1}^{T}  p(y_t|\hm{x}, \hm{y}_{<t}), \ T \geq 1
    \label{eq:fact}
\end{equation}
When $T=0$, $\hm{y}=(y_0)$ and $p(y_0| \hm{x})=1$ because $y_0$ is a fixed start token. Beam search \cite{graves2012sequence} is a decoding strategy to predict a target sequence by maximizing this factorization. Given a vocabulary set $\mathcal{V}$, at each inference step $t$, beam search selects a candidate beam set $\mathcal{B}^{\mathcal{K}}_{t} = \{\hm{b}_{t}^k\}_{k=1}^\mathcal{K}$ (where each beam $\hm{b}_{t}^k = (b_0^k,\cdots,b_t^k)$ is a candidate sequence) from an all-possible beam set $\mathcal{B}_{t}$ of size $\mathcal{K} \times |\mathcal{V}| $, namely,
\begin{equation}
    \mathcal{B}_t = \left\{\bm{b}_{t-1}^k \circ v\ |\ \bm{b}_{t-1}^k \in \mathcal{B}^{\mathcal{K}}_{t-1},\ v \in \mathcal{V}\right\}
\end{equation}

\begin{equation}
    \mathcal{B}^{\mathcal{K}}_t = \left\{\hm{b}^k_t\ |\ \hm{b}^k_t=\textrm{argtopk}\left(\log p(\hm{b}_t|\hm{x})\right),\ \hm{b}_t \in \mathcal{B}_t \right\},\ t \geq 1
    \label{eq:bt}
\end{equation}
where $\textrm{argtopk}(\cdot)$ outputs $\mathcal{K}$ beams with the highest conditional probability, and $\circ$ is the concatenation operation. Besides, $\mathcal{B}^{\mathcal{K}}_0=\{\hm{b}_{0}^k\}_{k=1}^\mathcal{K}$ where $b_0^k$ is the start token. By Eq.~\ref{eq:fact}, $\log p(\hm{b}_t|\hm{x})$ is an accumulated value. Its calculation can be simplified as:
\begin{equation}
    \log p(\hm{b}_t|\hm{x}) =\left\{
    \begin{aligned}
        &\log p(\hm{b}^k_{t-1}|\hm{x}) + \log p(v|\hm{x}, \hm{b}^k_{t-1}),\ &t \geq 2 \\
        & \log p(v|\hm{x}, b_0^k),\ &t=1
    \end{aligned}
    \right.
    \label{eq: accumu}
\end{equation}
where the value of $\log p(\hm{b}^k_{t-1}|\hm{x})$ is computed from the previous step. Therefore, at each step, we only need calculate the condition probability of each token in the vocabulary set. 

A beam is terminated after it generates an ending token, and the beam set of $\mathcal{K}$ terminated beams is defined as $\mathcal{Y}$. The final hypothesis $\hm{y}^*$ is chosen from $\mathcal{Y}$ based on the beam probability normalized by $length^a$ where $a$ is a hyper-parameter of length~\cite{Wu2016Google}:
\begin{equation}
    \hm{y}^* = \argmax_{\hm{y}^k \in \mathcal{Y}}\ \frac{\log p(\hm{y}^k\ |\ \hm{x})}{(|\hm{y}^k|-1)^{a}}
    \label{eq: beam-search}
\end{equation}
where $\hm{y}^k=(y_0^k,\cdots,y_T^k)$. $|\hm{y}^k|-1$ is used since the start token is not considered in calculating the length.

\subsection{Attention Distribution}
Attention distribution is a continuous vector whose element indicates the degree of attention paid to a source token. The element is formed by the accumulation of cross attention, i.e., $\sum_t\alpha_{t,i}$, where $\alpha_{t,i}$ refers to the cross attention that the $t_{th}$ token in the target sequence gives to the $i_{th}$ source token.\footnote{Mean pooling is used for multi-layers and multi-heads} Specially, cross attention is a scaled dot-product \cite{vaswani2017attention} of hidden states of the source $\hm{x}$ and the target sequence $\hm{y}$. Notably, since Transformer-decoder is an auto-regressive model, the cross attention assigned by $t_{th}$ target token is actually calculated by $\hm{y}_{<t}=(y_0, \cdots, y_{t-1})$.

{\bf Global Attention Distribution}. The global attention distribution $ \hm{g} = [g_1, \cdots,g_i,\cdots, g_n] \in \mathbb{R}^{n}$ is the attention distribution given by the reference, where global attention $g_i$ refers to the total attention that the reference attends to the $i_{th}$ source token, and $n$ is the source length. 

{\bf Optimal Length}. The summation of $ \hm{g}$, namely $\sum_{i=1}^n g_i$, is equal to $\sum_{i=1}^n \sum_{t=1}^T \alpha_{t,i}=T$ due to $\sum_{i=1}^n \alpha_{t,i} = 1$ in softmax operation, where $T$ is the reference length, or equivalently the optimal length $Z$. 

{\bf Local Attention Distribution}. The local attention distribution $ \hm{l}_{t}^k = [l_{t,1}^k,\cdots,l_{t,i}^k, \cdots,l_{t,n}^k] \in \mathbb{R}^{n}$ is the attention distribution of the $k_{th}$ generated sequence and updated at each decoding step $t$. Thereinto, the local attention $l_{t,i}^k$ denotes the total attention paid to the $i_{th}$ source token by the $k_{th}$ beam sequence $(b_1^k,\cdots,b_t^k)$ and is dependent on the sequence generated before $t$, i.e., $\hm{b}_{t-1}^k=(b_0^k,\cdots,b_{t-1}^k)$.

\section{Proposed Global-aware Inference} \label{Global-aware Inference}
\subsection{Global Scoring Mechanism}\label{sec:proposed}
The global scoring mechanism consists of an attention scoring function and a length reward function.
Given the global attention distribution $\hm{g}$, the attention scoring function $\mathcal{A}(\cdot)$ at the step $t$ depends on $\hm{b}_{t-1}^k$,
\begin{equation}
    \mathcal{A}(\hm{b}_{t-1}^k) =  \frac{\sum_{i=1}^n \min(l_{t,i}^k, g_i)}{\zeta_t^k} , \ \zeta_t^k= \sum_{i=1}^{n}{l_{t,i}^k},\ t \geq 1
    \label{eq:as}
\end{equation}
where $\zeta_t^k$ indicates the total attention that the generated sequence $(b_1^k,\cdots,b_t^k)$ gives to the source, and $\zeta_t^k = |\hm{b}_{t-1}^k|=t$ because the assignable attention for each generated token is $1$. Notably, Eq.~\ref{eq:as} attains the maximum score provided that each $l_{t,i} \leq g_i$. As mentioned in Section~\cref{sec:int}, the reason for this design is that we desire  the local attention $l_{T,i}$ at the termination is exactly $g_i$, since the final hypothesis is expected to attend to source tokens in the global-optimal manner. Meanwhile, we have $l_{T,i} > l_{t,i}$ for $t < T$ because $l_{t,i}$ monotonically increases on $t$ with $\alpha^l_{m,i} > 0$. Therefore, at any step, the local attention $l_{t,i}$ should not surpass $g_i$. Otherwise, the attention score will decline, and the penalty depends on the total amount by which these $l_{t,i}$ exceed $g_{i}$ (see Theorem \ref{th:1}). Further, the attention score could be considered as the proportion of correctly assigned attention to the total attention given by the generated sequence, where correct assignment indicates that all parts of $l_{t,i}$ do not exceed $g_{i}$. Also, it could be interpreted as the correct allocation probability of local attention against the global attention distribution (see Corollary~\ref{th:2}). In this case, the total attention score can be expressed as the same multiplicative form as Eq.~\ref{eq:fact} to be elegantly integrated into beam search.
\begin{customthm}{1}\label{th:1}
Let $\mathcal{M} = \left\{s\ :\ l_{t,s}^k > g_s \right\}$ and $\Delta_t^k = \left\{\delta\ |\ \forall s\in \mathcal{M},\ \delta=l_{t,s}^k-g_s\right\}$. Given $\Delta = \sum_{\delta\in\Delta_t^k}\delta$ where $\Delta \geq 0$, then $\mathcal{A}(\hm{b}_{t-1}^k)$ decreases as $\Delta$ increases.
\end{customthm}

\begin{customcoro}{1.1}\label{th:2}
The bound of $\mathcal{A}(\hm{b}_{t-1}^k)$ is between $0$ and $1$.
\end{customcoro}

\begin{proof}
See App.~\ref{app: pf}.
\end{proof}

In addition to the constraint for $l_{t,i}$, we still desire $l_{T,i} = g_i$ for each token. An ideal hypothesis should have two characteristics simultaneously. Namely, its attention score at the termination is the maximum $1$, and its length equals the optimal length. Therefore, we introduce a length reward function to cooperate with the attention score to penalize the situation $l_{T,i} \neq g_i$, which will be discussed at the end of this subsection.

As mentioned before, the total attention score at the decoding step $t$ is defined as:
\begin{equation}
     A(\hm{b}^k_{t-1}) = \prod_{m=1}^{t} \mathcal{A}(\hm{b}^k_{m-1})
\end{equation}
Thus, the joint scoring function $J(\cdot)$ is modified from Eq.~\ref{eq: accumu}:
\begin{equation}
    \begin{aligned}
    J(\hm{b}_t,\hm{x}) 
     & = \log p(\hm{b}^k_{t-1}|\hm{x}) +\beta \log A(\hm{b}^k_{t-1})+ \log p(v|\hm{x}, \hm{b}^k_{t-1}) \\
     & = \sum_{m=1}^{t-1} \left(\log p(b_m|\hm{x}, (\hm{b}^k_{t-1})_{<m}) + \beta \log \mathcal{A}(\hm{b}^k_{m-1})\right) +\log p(v|\hm{x}, \hm{b}^k_{t-1})+\beta \log \mathcal{A}(\hm{b}^k_{t-1})\\
     & = J(\hm{b}^k_{t-1},\hm{x})  +\log p(v|\hm{x}, \hm{b}^k_{t-1})+\beta \log \mathcal{A}(\hm{b}^k_{t-1}),\ t \geq 2 \\
\end{aligned}
\label{eq:joint}
\end{equation}
and $J(\hm{b}_1,\hm{x}) = \log p(v|\hm{x}, b^k_0)+\beta \log \mathcal{A}(b^k_0)$, where $\beta$ is a hyper-parameter to  trade-off between the probability and attention score. Similar to $\log p(\hm{b}_t|\hm{x}) $ in Eq.~\ref{eq: accumu}, $J(\hm{b}_t,\hm{x})$ is also an accumulative score. Consequently, at each step $t$, we only need  compute $ p(v|\hm{x}, \hm{b}^k_{t-1})$ and $\mathcal{A}(\hm{b}^k_{t-1})$. Compared with Eq.~\ref{eq: accumu}, the time complexity of each step is only increased by $\mathcal{O}(\mathcal{K})$ as there are $\mathcal{K}$ attention scores. Replacing $\log p(\hm{b}_t|\hm{x})$ in Eq.~\ref{eq:bt} by $J(\hm{b}_t,\hm{x})$, we can select the top $\mathcal{K}$ beams of each decoding step according to not only the probability distribution conditional on local information $\hm{b}^k_{t-1}$ but also the score conditional on global information $\hm{g}$.

Considering the length reward function, the final hypothesis is thus defined as:
\begin{equation}
    \hm{y^*} = \argmax_{\hm{y}^k \in \mathcal{Y}}\left(\ \frac{J(\hm{y}^k,\hm{x})}{|\hm{y}^k|-1} + \beta \gamma R(\zeta_T^k, Z)\right)
    \label{eq:final}
\end{equation}
where $R(\cdot)$ is the length reward function dependent on the optimal length $Z$ and the candidate hypothesis length $\zeta_T^k$. Exactly, $\zeta_T^k$ is the total attention that a candidate hypothesis $(y_1^k, \cdots, y_T^k)$ pays to the source and equals $|\hm{y}^k|-1$. Besides, the attention score and length reward are weighted by a hyper-parameter $\gamma$, and the role of $\beta$ is to ensure that the two are at the same level relative to the probability. We remove $a$ in Eq.~\ref{eq: beam-search} as it only adjusts the length preference without really controlling the length. 

The design of $R(\cdot)$ could be straightforward -- one only need ensure that it  increases as $\zeta_T^k$ approaches $Z$, and reaches the maximum only at $\zeta_T^k=Z$. In this paper, we design a step-wise length reward function $\mathcal{R}(\zeta_t^k, Z)$ to better balance the relationship between the attention score and the length reward and make the whole searching process as succinct as beam search. We put the design details of the step-wise length reward in App.~\ref{app: lr}, and we regard Eq.~\ref{eq:final} as the general scoring formulation of global-aware inference.

\subsection{Predict the Global Attention Distribution}
Since the reference is unknown practically, the global attention distribution could only be predicted from the source. We construct an attention-prediction model to learn the relationship between the source tokens and the global attention distribution.

The input of the attention-prediction model is the fixed encoder output $\hm{E} \in \mathbb{R}^{n\times d}$ of BART or PEGASUS plus
learnable positional encodings $\hm{P} \in \mathbb{R}^{n\times d}$, where $d$ is the dimension of hidden states. The input is fed to a learnable Transformer-encoder to obtain $\widetilde{\hm{E}}\in \mathbb{R}^{n\times d}$ that is encoded with additional context information, followed by a linear transformation with an exponential function:
\begin{equation}
    \widehat{\hm{g}} = \exp\left(\widetilde{\hm{E}}\hm{W}_g  + \hm{b}_g\right)
\end{equation}
where $\widehat{\hm{g}}\in \mathbb{R}^{n}$ refers to the prediction of $\hm{g}$, $\hm{W}_g \in \mathbb{R}^{d \times 1}$ and $\hm{b}_g\in \mathbb{R}^{n}$ are the learnable weights and biases. The exponential function is imposed to ensure $ \widehat{\hm{g}} > 0$. We choose the exponential function for this operation because it is shown stable in the training and testing stage. Given the objective of minimizing the distance between $ \widehat{\hm{g}}$ and $\hm{g}$, the loss is defined as their Euclidean distance:
\begin{equation}
    \mathcal{L} = \Vert \widehat{\hm{g}}-\hm{g} \Vert_2
\end{equation}
The predicted optimal length $\widehat{Z}$ is the sum of elements in $\widehat{\hm{g}}$. Note that the length reward function is not affected no matter whether $\widehat{Z}$ is an integer or not.

\section{Experiment}\label{sec: exper}
\subsection{Setup} \label{sec: setup}
{\bf Datasets}. We evaluate the performance on totally $9$ summarization datasets, where $2$ datasets (CNN/DM \cite{teaching2015}, XSUM \cite{dong2019unified}) with BART \cite{lewis-etal-2020-bart} and $8$ datasets (XSUM \cite{dong2019unified}, BillSum \cite{kornilova-eidelman-2019-billsum}, Multi-News \cite{fabbri-etal-2019-multi}, NewsRoom \cite{grusky-etal-2018-newsroom},  WikiHow \cite{koupaee2018wikihow}, Reddit TIFU \cite{volske-etal-2017-tl}, arXiv and PubMed \cite{Cohan_2018}) with PEGASUS \cite{Zhang2020PEGASUSPW}. Thereinto, XSUM \cite{dong2019unified} is a highly abstractive dataset whose summaries are all expressed in a short sentence. 

{\bf Implementation Details}.
We adopt a randomly initialized 2-layer transformer-encoder in the attention-prediction model wherethe structure of each layer is the same as the BART-encoder layer. The optimizer is the Adabelief-optimizer~\cite{zhuang2020adabelief} with eps $1e-16$, betas $(0.9, 0.999)$, weight decay $1e-4$ and learning rate $2e-5$. The attention-prediction model is trained on the training set for about $50,000$ steps, and checkpoints are saved per $10,000$ steps to select the best checkpoints on the development set. Since the attention prediction is slightly different from common sequence tagging tasks, we have summarized two notable points after several attempts -- the dropout rate should be $0$, and a small learning rate is preferred. All experiments are conducted on $3$ {\it RTX 6000}. We include the global-aware inference in the generation code of {\it HuggingFace transformers} \cite{wolf-etal-2020-transformers}. At the time of evaluation, ROUGE-1, ROUGE-2 \& ROUGE-L (R-1, R-2 \& R-L) scores \cite{lin-2004-rouge} are computed from the ROUGE code\footnote{https://github.com/pltrdy/files2rouge} used by BART \cite{lewis-etal-2020-bart}.

{\bf Hyper-parameter Selection}. Although the global-aware inference requires two new hyper-parameters $\gamma$ and $\beta$, some original hyper-parameters of beam search, namely length penalty, minimum and maximum length, are omitted. The searching scopes of $\beta$ and $\gamma$ are in $\{2, 4, 6, 10, 12, 15, 18, 20\}$ and $\{0, 0.5, 1, 1.5, 2\}$, respectively. According to the numerical tests on the development set, we finally choose $\beta=12,\ \gamma=1.5$ for CNN/DM and $\beta=4,\ \gamma=0$ for XSUM. As limited improvement could be observed from larger $\gamma$'s, we recommend $\gamma=1$ for normal or longer targets. When testing the global-aware inference with PEGASUS \cite{Zhang2020PEGASUSPW}, we directly use {\bf empirical hyper-parameters} for each dataset, namely $\beta=4,\ \gamma=0$ for XSUM and $\beta=12,\ \gamma=1$ for other $7$ datasets. The beam size $\mathcal{K}$ follows the setups in BART \cite{lewis-etal-2020-bart} and PEGASUS \cite{Zhang2020PEGASUSPW}.

\subsection{Results}\label{sec: results}
\begin{table*}[htbp]\scriptsize
   \centering
   \caption{ROUGE $F_1$ scores of summaries generated by global-aware, in comparison to beam search with length regularizations. Notably, global-aware uses empirical hyper-parameters.}

    \begin{tabular}{lccc|ccc|ccc|ccc}
    \toprule
     &\multicolumn{3}{c}{\bf XSUM } &\multicolumn{3}{c}{\bf BillSum } &\multicolumn{3}{c}{\bf Multi-News}&\multicolumn{3}{c}{\bf WikiHow}\\
     $\mathcal{K}=8$& {\bf R-1} & {\bf R-2} & {\bf R-L}  & {\bf R-1} & {\bf R-2} & {\bf R-L} & {\bf R-1} & {\bf R-2} & {\bf R-L}& {\bf R-1} & {\bf R-2} & {\bf R-L}\\ 
   
    \midrule
    {\bf Beam search} \cite{Zhang2020PEGASUSPW} & 47.05 & 24.53 & 39.33 & 57.00 & 39.65 & 52.70& 47.29 & 18.91 & 43.31 & 41.86 & 19.04 & 40.40 \\
   
       {\bf Global-aware} & {\bf 47.33} & {\bf 24.66} & {\bf 39.50} & {\bf 58.66} & {\bf 40.12 } & {\bf 53.96}& {\bf 47.95} & {\bf 19.08} & {\bf 43.93}& {\bf 42.82} & {\bf 19.68} & {\bf 41.43}\\
       \midrule \midrule

       &\multicolumn{3}{c}{\bf Reddit TIFU } &\multicolumn{3}{c}{\bf NewsRoom } &\multicolumn{3}{c}{\bf PubMed}&\multicolumn{3}{c}{\bf arXiv}\\ 
        & {\bf R-1} & {\bf R-2} & {\bf R-L}  & {\bf R-1} & {\bf R-2} & {\bf R-L} & {\bf R-1} & {\bf R-2} & {\bf R-L}& {\bf R-1} & {\bf R-2} & {\bf R-L}\\ \midrule
       
    {\bf Beam search} \cite{Zhang2020PEGASUSPW} & 27.55 & 8.67 & 22.12 & 42.05 & 29.88 & 38.70& 44.25 & 19.19 & 41.11 & 43.82 & 16.75 & 39.28 \\
   
       {\bf Global-aware} & {\bf 28.31} & {\bf 9.13} & {\bf 23.30} & {\bf 44.68} & {\bf 31.71 } & {\bf 41.28}& {\bf 45.78} & {\bf 20.16} & {\bf 42.62}& {\bf 44.92} & {\bf 17.41} & {\bf 40.31}\\
       \bottomrule
    \end{tabular}
     \label{tab:peg-rouge}
\end{table*}
{\bf Comparison with Beam Search}. Beam search is a hard-to-beat baseline which has stood the test of time and proven its superiority in practice for long \cite{meister2020if}. In Table \ref{tab:peg-rouge}, we compare our global-aware inference to beam search with length regularizations (i.e., $\alpha$ in Eq.~\ref{eq: beam-search}, accompanied with two hard constraints, namely minimum length and maximum length). We strictly follow the hyper-parameter setups of PEGASUS \cite{Zhang2020PEGASUSPW} in terms of beam search, while we only adopt empirical hyper-parameters for our method. Even so, significant improvements can be observed on all the data sets, especially when the summary is of normal or longer length.

\begin{table*}[htbp]\scriptsize
\centering
 \begin{minipage}[t]{0.44\textwidth}
       \caption{Comparison  with other methods.}
    \begin{tabular}{lccc}
    \toprule
    &\multicolumn{3}{c}{\bf CNN/DM} \\
     $\mathcal{K}=4$& {\bf R-1} & {\bf R-2} & {\bf R-L} \\ \midrule
       {\bf Beam search} \cite{lewis-etal-2020-bart} & 44.12 & 21.21 &  40.89\\
       {\bf  + Our coverage} & 44.74 & 21.69 &  41.48  \\
       {\bf + Repetition penalty} \cite{keskar2019ctrl} & 44.11 & 21.14 &  40.87 \\
       {\bf + Attention masking} \cite{cao2021attention} & {\bf 45.54} & {\bf 22.24} &  {\bf 42.44} \\
          \midrule
      {\bf Global-aware}& 45.13&  21.77 &  42.04 \\
      \midrule\midrule
      
       &\multicolumn{3}{c}{\bf XSUM} \\
         $\mathcal{K}=6$& {\bf R-1} & {\bf R-2} & {\bf R-L} \\ \midrule
    
       {\bf Beam search} \cite{lewis-etal-2020-bart} & 45.38 & 22.32 & 37.15\\
       {\bf + Our coverage}  & 44.54 & 21.82 & 36.97 \\
       {\bf + Repetition penalty} \cite{keskar2019ctrl} & 45.40 & 22.31 & 37.13\\
       {\bf + Attention masking} \cite{cao2021attention}& 45.35 & 22.31 & 37.15\\
          \midrule
      {\bf Global-aware}& {\bf 45.57} & {\bf 22.60} & {\bf 37.61}\\

       \bottomrule
    \end{tabular}
     \label{tab:compare}
      \end{minipage}
       \hspace{0.5in}
 \begin{minipage}[t]{0.42\textwidth}
   \caption{ORACLE  and ablation results.}
    \begin{tabular}{lccc}
    \toprule
     &\multicolumn{3}{c}{\bf CNN/DM} \\
    & {\bf R-1} & {\bf R-2} & {\bf R-L}  \\ \midrule
      {\bf ORACLE global-aware }  & {\bf 51.85} & {\bf 28.13} &   {\bf 48.68}  \\
        {\it -w/o length reward} & 50.46 & 27.53 &  47.43 \\\midrule
        {\bf Global-aware} & {\bf 45.13}&  {\bf 21.77} & {\bf  42.04}  \\
        {\it -w/o length reward} & 44.39 & 21.58 &  41.41\\
        {\it -w/o attention score} & 44.12 & 21.29 &  40.91\\ \midrule
        \midrule
         &\multicolumn{3}{c}{\bf XSUM} \\
     & {\bf R-1} & {\bf R-2} & {\bf R-L}  \\ \midrule
        {\bf ORACLE global-aware}  & {\bf 49.50} & 26.24 & 41.13  \\
        {\it -w/o length reward} & 48.92 & {\bf 26.48} & {\bf 41.45}  \\\midrule
        {\bf Global-aware} ($\gamma=1$) & 45.44 & 22.15 & 37.11  \\
        {\it -w/o length reward} &{\bf 45.57} & {\bf 22.60} & {\bf 37.61}\\
        {\it -w/o attention score} & 45.23 & 21.88 &  36.73\\
       \bottomrule
    \end{tabular}
     \label{tab:oracle}
     \end{minipage}
\end{table*}

\begin{table*}[htbp]\scriptsize
   \centering
   \caption{Improvements of attention head masking and global-aware on beam search \cite{Zhang2020PEGASUSPW} in terms of ROUGE-L $F_1$ score. Both use empirical setups.}

    \begin{tabular}{lcccccccc}
    \toprule
     & {\bf XSUM}   & {\bf BillSum} & {\bf Multi-News} & {\bf WikiHow} & {\bf Reddit}& {\bf NewsRoom} & {\bf PubMed} & {\bf arXiv}\\
    \midrule
    {\bf Attention head masking} \cite{cao2021attention} & -0.31  & 0.23 & 0.34 & 0.10 & -0.16 & 1.24 & 0.35 & 0.21\\

    {\bf Global-aware} & {\bf 0.17}  & {\bf 1.26} & {\bf 0.62} & {\bf 1.03} & {\bf 1.18} & {\bf 2.58} & {\bf 1.51} & {\bf 1.03}\\
    \bottomrule
    \end{tabular}
     \label{tab:im}
\end{table*}

{\bf Comparison with Other Attention Approaches}.
In this part, we focus on comparing other approaches which also exploit certain attention distributions to improve beam search. The first is the coverage penalty~\cite{Wu2016Google, li2018simple}. To enhance its performance in summarization, we replace its preset attention distribution with our predicted global attention distribution. Note that the coverage function can only evaluate the generated sentences at the terminal step. Instead of comparing the global-aware inference to the methods~\cite{See2017, li-etal-2018-improving-neural, gehrmann-etal-2018-bottom} that aim to reduce repetition using the dynamic attention distribution, we compare our algorithm with the CTRL repetition penalty~\cite{keskar2019ctrl} which has similar motivation but is more systematical and independent of training.  Table~\ref{tab:compare} lists the comparison results against different algorithms. It can be observed that our global-aware approach can improve the performance of beam search stably and fundamentally. We also observe that the attention head masking~\cite{cao2021attention} appears to outperform the global-aware approach on CNN/DM, but it fails to gain any improvement on XSUM. To further show the advantage of the proposed approach, we will take a closer examination on  the attention head masking~\cite{cao2021attention} and our proposed approach in the next part.

{\bf Further Comparison with Attention Head Masking}.
First, one should bear in mind that the attention head masking~\cite{cao2021attention} acts on the model instead of beam search, which is opposite to us. Specifically, it selects contents during decoding by disabling partial attention heads for unimportant tokens to decrease the attention to these tokens. According to the reported results presented in Table \ref{tab:compare}, we can see that although attention masking achieves amazing results on CNN/DM, it does fail completely on XSUM.  Since hiding unimportant tokens from some heads results in the loss of context information of salient tokens, this would lead to its instability. Thus, it could be ineffective for tasks that require contextual information of the whole source such as XSUM. Taking a further comparison, we deploy the attention head combinations selected for CNN/DM and XSUM to examine its effect on PEGASUS~\cite{Zhang2020PEGASUSPW}. These comparison results are shown in Table~\ref{tab:im}. Evidently, our method enjoys a robust feature that is able to boost summary inference on various datasets and models even with the same set of hyper-parameters. In contrast, attention masking~\cite{cao2021attention} behaves much sensitive to the changes of models and datasets. Besides, attention masking has to construct its training saliency labels based on the longest common subsequences between a reference summary and the source. This may be hardly achieved in some text generation tasks (e.g. translation) where no common subsequence exists at all. Such drawback presents one main limitation for attention masking. 

{\bf ORACLE Results and Ablation Study}. ORACLE refers to the global-aware inference combined with (true) global attention distribution instead of predicted global attention distribution. The related results have been presented in Table \ref{tab:oracle}, and the significant boosting certifies that the proposed method could improve beam search with the global attention distribution. On the other hand, we conduct ablation experiments on ORACLE and global-aware. Both results indicate that length reward plays an important role in generating normal-length text but causes adverse effect on generating text of very short length. Besides, the performance declines significantly when only length reward is applied, due to the fact that sole length reward cannot calibrate beam search step-wise.

\begin{figure}
\begin{minipage}[t]{0.45\textwidth}
    \begin{minipage}[t]{0.98\textwidth}
    \centering
    \scriptsize
     \captionof{table}{Generate CNN/DM summaries with XSUM's style.}
    \begin{tabular}{lcc}
    \toprule
    {\bf R1/R2/RL}&{\bf Shorter beam search} & {\bf Global-aware}$\dagger$\\
    \midrule
    {\bf $F_1$ score}& {\bf 43.6}/{\bf 20.9}/40.4 &  {\bf 43.6}/20.4/{\bf 40.6} \\
     {\bf Recall}&{\bf 48.9/23.4/45.2}&40.4/18.8/37.6\\
      {\bf Precision}&41.4/19.9/38.3&{\bf 50.5/23.8/47.0}\\
       \bottomrule
    \end{tabular}
     \label{tab:cross}
    \end{minipage}
   
     \begin{minipage}[t]{0.98\textwidth}
     \scriptsize
     \centering
    \captionof{table}{Generate summaries with corrupted attention distributions.}
    \begin{tabular}{lccc}
    \toprule
    &{\bf R-1}& {\bf R-2} & {\bf R-L}\\
    \midrule
    {\bf Beam search}&27.00&12.21&23.88\\
    {\bf Global-aware (10K)}&28.78&12.82&25.51\\
     {\bf  Global-aware (100K)}&{\bf29.59}&{\bf13.72}&{\bf26.30}\\
       \bottomrule
    \end{tabular}
     \label{tab:domain}
    \end{minipage}
    \end{minipage}
\begin{minipage}[t]{0.54\textwidth}
    \centering
    \caption{Sensitive analysis in the test set.}
    \includegraphics[width=5.8cm]{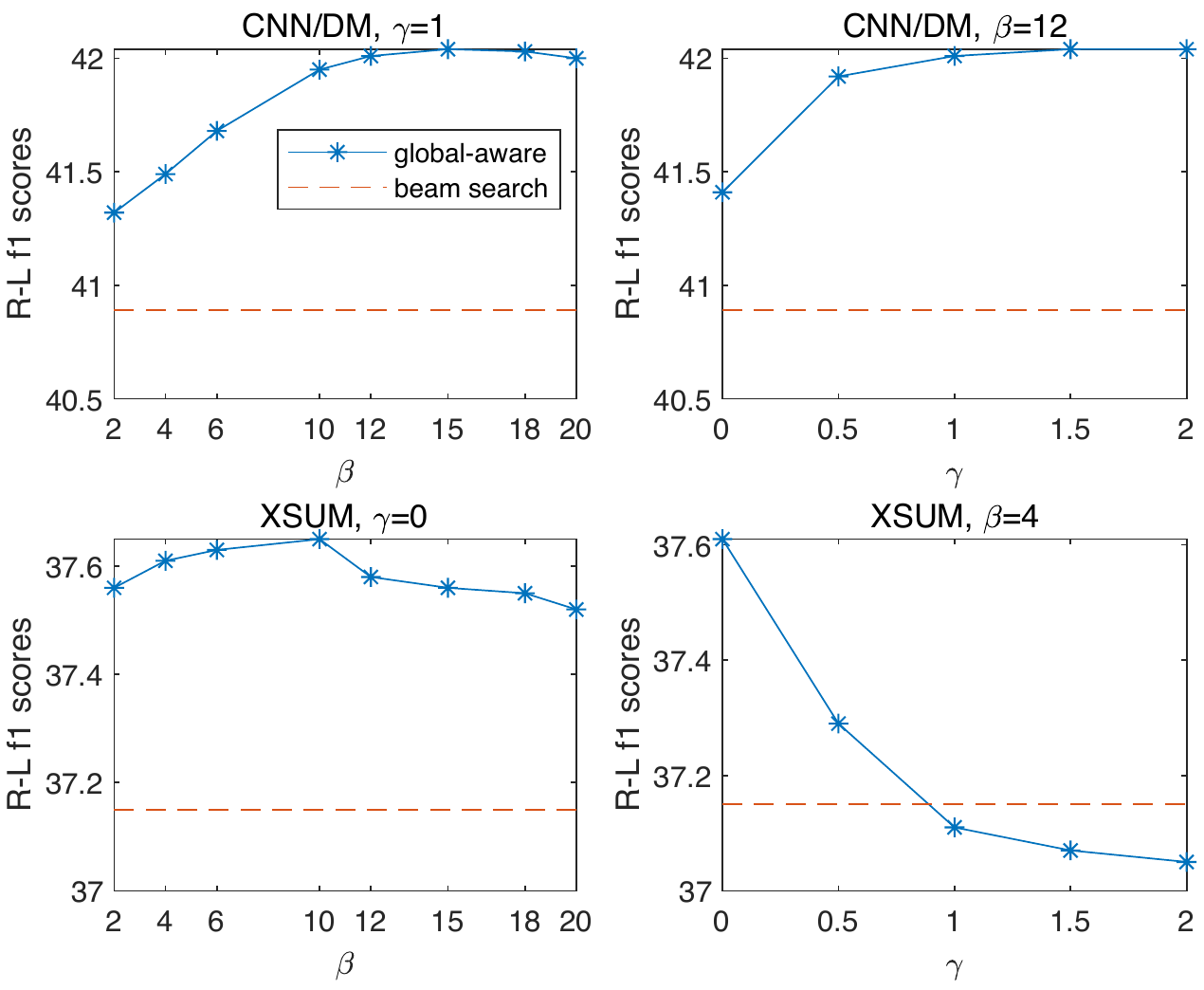}
    \label{fig:sensitive}
    \end{minipage}
\end{figure}

 {\bf Robustness}. We intend to examine the robustness of our proposed global-aware algorithm in this and next part. To do so, we substitute the parameters in CNN/DM's attention prediction model to XSUM's to create more abstractive summaries for CNN/DM. For comparison, we set the minimum length of beam search as $0$ to allow it to generate shorter summaries. Table~\ref{tab:cross} shows the $F_1$ score, recall and precision of the shorter beam search and global-aware$\dagger$. It is surprising that even by using the attention distribution from a different dataset with distinct properties, the proposed global-aware mechanism still manages to generate meaningful summaries with competitive $F_1$ scores, proving the robustness of this algorithm. Moreover, the higher Precision and lower Recall of the global-aware suggest that although information is partially lost, the algorithm still summarizes core information in a concise format, compared to the standard beam search. On the other hand, we exploit a BART model fine-tuned on CNN/DM to generate summaries of NewsRoom directly, and the ROUGE scores of beam search are shown in Table~\ref{tab:domain}. Next, we randomly select 10K and 100K samples from the training set and use them to fine-tune the attention-prediction model, where the global-aware improves beam search substantially. The experiment once again validates the robustness of the proposed inference algorithm, as it maintains a reasonably good performance even from learning a corrupted attention distribution from a BART without fine-tuning.

{\bf Sensitive Analysis}. We further examine the performance robustness with respect to the hyper-parameters. From Figure \ref{fig:sensitive}, we could see the global-aware inference is always better than beam search in CNN/DM, no matter how its hyper-parameters change. Besides, the performance is less sensitive to the hyper-parameters when $\beta \geq 10$ or $\gamma \geq 1$. While in XSUM, the global-aware could improve beam search stably with $\gamma=0$, but there is a significant decline when applying length reward. In fact, the attention score favors shorter hypotheses, and the length reward could alleviate the bias. However, if the references of a dataset are already very short such as XSUM, the length reward may lead to a counterproductive effect. Since the setup of CNN/DM is applicable to most datasets, we argue that the global-aware inference is robust to both hyper-parameters in most cases.

\section{Analysis on Global Attention Distribution}

\subsection{Distribution Law}
\begin{figure}
\centering
\includegraphics[scale=0.65]{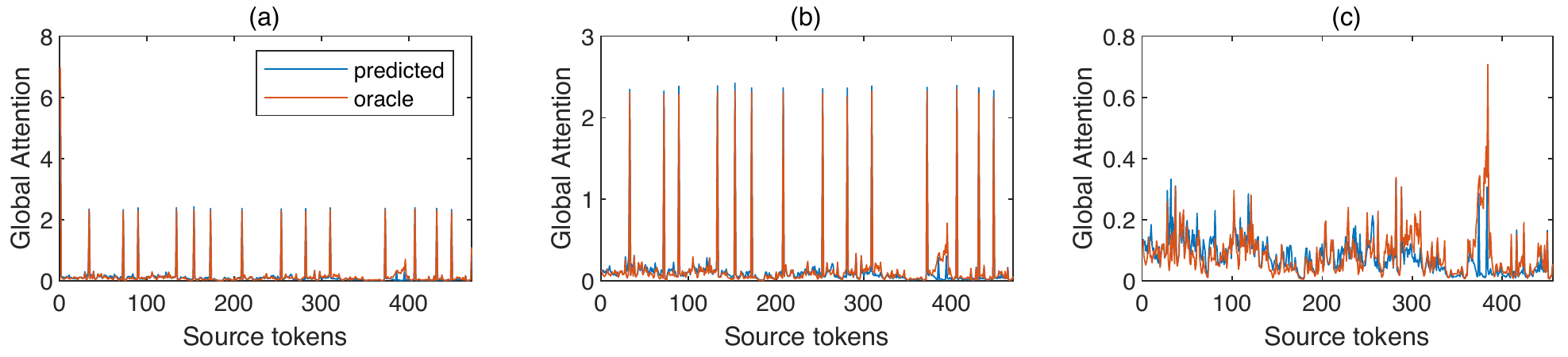}
\caption{Predicted and ORACLE global attention in BART. There are attention distributions of (a) the whole source, (b) the source without the start \& end tokens, (c) the source without the start \& end tokens and full stops.}
\label{fig:ga}
\end{figure}
The distribution law of global attention in BART~\cite{lewis-etal-2020-bart} is shown in Figure \ref{fig:ga}. It is observed that most attention is assigned to the start token and full stops which are not semantically salient, and most of the remaining attention is unevenly allocated to (semantically) overlapped tokens between the source and the reference (i.e., important words). It is worth mentioning that the importance here is no longer a simple binary classification like in \cite{gehrmann-etal-2018-bottom,cao2021attention}, but a continuous numerical value decided by the model knowledge learned from data. In general, one should not simply equate the global attention with the word importance, but should be clear that it essentially reflects knowledge learned by the attention model such as the syntactic and semantic structure of sentences.
Meanwhile, the distribution law indicates that attention distributions in pre-trained models may not be relevant with the uniform distribution at all. That is to say, it is not reasonable to still use an uniform attention threshold (like the threshold $1$ preset in \cite{Wu2016Google, gehrmann-etal-2018-bottom}) to regulate the decoding of pre-trained models. Last but not least, our general motivation is to alleviate locality bias by aligning the attention distribution of reference and hypothesis, which does not really care how the global attention is distributed only if it is predictable. However, the proposed penalty mechanism is indeed insensitive to some distributions, and we will provide more thinking about this in App.~\ref{app:nmt} by applying the  global-aware inference to translation.

\subsection{ Why It can be Predicted from Source?}
\begin{figure}
\centering
\includegraphics[scale=0.65]{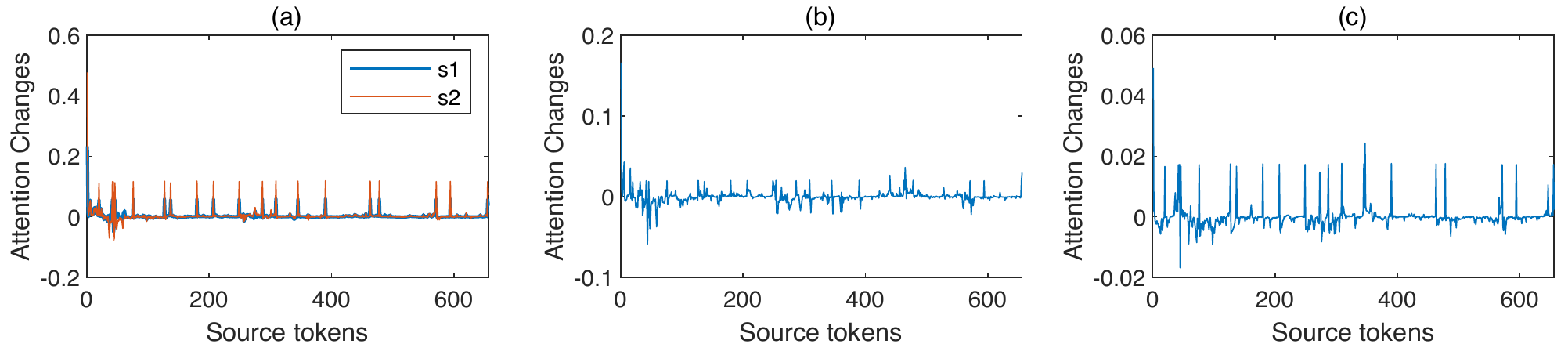}
\caption{Changes of the attention distribution when (a) one word in the reference is replaced by a similar word (s1) and a random word (s2), (b) the sentence order of the reference is shuffled, (c) a factual knowledge in the reference is distorted.}
\label{fig:attc}
\end{figure}
Since the ORACLE experiments indicated that the global attention is helpful for decoding, the concern remains  if it is predictable using only source. In App.~\ref{app: dp} we will show the predictability empirically, while here we just provide an interesting explanation. In our opinion, the global attention distribution is an interpretable representation of reference which not only has the characteristics of hidden representation but also can be explained by the source tokens. First of all, given the source, the global attention is calculated by the input reference and trained neural network parameters; this is similar to achieving hidden representation. Moreover, like hidden representation, the global attention distribution could also capture the semantic similarity, e.g., replacing a reference word with a semantically similar word leads to slighter attention changes than that with a random word (see Figure \ref{fig:attc}.a). Besides, it is observed from Figure \ref{fig:attc}.b and Figure \ref{fig:attc}.c that global attention is able to represent the sentence order and factual knowledge. On the other hand, the global attention distribution enjoys a distinct characteristic that each  of its features can be explained as the degree of attention paid to a source token, which means changes of such representation are regular to some extent. For example, in Figure \ref{fig:attc}.c,  we distort a numerical number in the reference to violate the fact stated in the source, and then find that the attention assigned to the actual number originally is most transferred to the special tokens and punctuation. Overall, similar sources should have similar global attention distributions, since  similar sources often have similar references and global attention distribution is a representation of reference. Moreover, the global attention and source tokens are in an one-to-one correspondence. Thereby, we argue that it is convenient to use the source to predict the global attention distribution.

\section{Related Work}
In the field of text generation, efforts have been made to boost beam search with regards to the attention distribution. For instance, some studies engage the attention distribution to penalize the situation where sources fail to be fully covered in translation tasks \cite{Wu2016Google, li2018simple}, while others \cite{gehrmann-etal-2018-bottom,See2017,li-etal-2018-improving-neural} incorporate dynamic attention distributions to evade tokens that have been highly regarded to reduce repetition. However, none of the aforementioned studies attempts to apply the global attention distribution to acquire the knowledge that the level of attention should be allocated to each token from the reference. Further, the existing score functions used by those studies are rather different from the proposed global scoring mechanism. More details can be seen in App.~\ref{app: smc}.

In addition to the attention distribution, other techniques are developed to improve beam search in terms of length bias~\cite{yang-etal-2018-breaking}, diversity-less~\cite{vijayakumar2018diverse}, vapidity~\cite{holtzman2019curious} and degradation~\cite{ott2018analyzing, cohen2019empirical}. 
These methods are not included for comparison because they are not suitable for summarization~\cite{yang-etal-2018-breaking, holtzman2019curious}, or do not aim to enhance beam search as the main purpose~\cite{vijayakumar2018diverse, ott2018analyzing,cohen2019empirical}. Besides, we argue that these patches to beam search are supposed to be hard for improving the performance stably and fundamentally (as shown in Table~\ref{tab:compare} given by our global-aware method) because they fail to specify what a final hypothesis should look like and are easy to trap into local optimums.

\section{Conclusion}
Beam search tends to fall into local optimums due to the lack of global information during inference. To calibrate beam search on the premise of global awareness, this study proposes a predictable protocol to stipulate how a global optimal hypothesis should attend to source tokens. By training a simple prediction model of the global attention distribution, a novel global scoring mechanism is then developed to regulate the inference on a step-wise basis. Our experiment shows that the proposed global-aware beam search generates higher-quality summaries, relative to beam search and other counterparts of the similar nature. Besides, the proposed algorithm is proven robust in generating meaningful texts even with corrupted attention distributions, implying its potential to cooperate with user-defined global attention distributions. We plan to focus our future study on generalizing the global-aware inference to a broader range of text generation tasks, including not only text2text but also image caption~\cite{Vinyals_2015_CVPR}, multimodal summarization~\cite{li2020vmsmo}, graph2text \cite{koncel2019text} and data2text~\cite{puduppully2019data}. 

\begin{ack}
Funding in direct support of this work: XJTLU Key Programme Special Fund  KSF-A-14 and KSF-P-02, National Natural Science Foundation of China under No. 61876155 and Jiangsu Science and Technology Programme under No. BE2020006-4.
\end{ack}

\bibliographystyle{plain}
\bibliography{att}

\clearpage
\appendix

\section{Proof}\label{app: pf}
\begin{customthm}{1}
Let $\mathcal{M} = \left\{s\ :\ l_{t,s}^k > g_s \right\}$ and $\Delta_t^k = \left\{\delta\ |\ \forall s\in \mathcal{M},\ \delta=l_{t,s}^k-g_s\right\}$. Given $\Delta = \sum_{\delta\in\Delta_t^k}\delta$ where $\Delta \geq 0$, then $\mathcal{A}(\hm{b}_{t-1}^k)$ decreases as $\Delta$ increases.
\end{customthm}

\begin{proof}
By the formula derivation, Eq.~\ref{eq:as} can be converted to:
\begin{equation}
\begin{aligned}
      \mathcal{A}(\hm{b}_{t-1}^k) & =  \frac{\sum_{i=1}^n \min(l_{t,i}^k, g_i)}{\zeta_t^k} \\
      & = \sum_{i=1}^n \min \left(\frac{l_{t,i}^k}{\zeta_t^k}, \frac{g_i}{\zeta_t^k}\right)\\
      & = \sum_{i=1}^n \min \left(0, \frac{g_i-l_{t,i}^k}{\zeta_t^k}\right) + \sum_{i=1}^n \frac{l_{t,i}^k}{\zeta_t^k}\\
      \label{eq:der1}
\end{aligned}
\end{equation}
Since $\zeta_t^k= \sum_{i=1}^{n}{l_{t,i}^k}$, we have $ \sum_{i=1}^n \frac{l_{t,i}^k}{\zeta_t^k} = 1$. Besides, with $\zeta_t^k>0$,
\begin{equation}
    \forall s \in \mathcal{M},\  \min \left(0, \frac{g_s-l_{t,s}^k}{\zeta_t^k}\right) = \frac{g_s-l_{t,s}^k}{\zeta_t^k} = \frac{-\delta}{\zeta_t^k} 
\end{equation}
Thus Eq.~\ref{eq:der1} is equal to:
\begin{equation}
\begin{aligned}
     \mathcal{A}(\hm{b}_{t-1}^k) &=0+ \sum_{\delta\in\Delta_t^k} \frac{-\delta}{\zeta_t^k} +1\\
     &=1 - \frac{\Delta}{\zeta^k_t}
\end{aligned}
\end{equation}
Then it is easy to prove that $\mathcal{A}(\hm{b}_{t-1}^k)$ goes down as $\Delta$ goes up.
\end{proof}

\begin{customcoro}{1.1}
The bound of $\mathcal{A}(\hm{b}_{t-1}^k)$ is between $0$ and $1$.
\end{customcoro}

\begin{proof}
Since we have achieved $\mathcal{A}(\hm{b}_{t-1}^k)=1 - \frac{\Delta}{\zeta^k_t}$ where $\mathcal{A}(\hm{b}_{t-1}^k)$ decreases monotonically on $\Delta$ and $\Delta\geq 0$, $\mathcal{A}(\hm{b}_{t-1}^k)$ reaches the maximum $1$ when $\Delta=0$. Next, we assume a beam can be generated indefinitely. In this case, there must be a situation that each $l_{t,i}^k > g_i$. Therefore we have
\begin{equation}
\begin{aligned}
     \mathcal{A}(\hm{b}_{t-1}^k) &=1-\frac{\sum_{i=1}^n (l^k_{t,i}-g_i)}{\zeta^k_t}\\
     &= 1-\frac{\sum_{i=1}^n l^k_{t,i}- \sum_{i=1}^n g_i}{\zeta^k_t}\\
     &=1-\frac{\zeta^k_t- \sum_{i=1}^n g_i}{\zeta^k_t} \\ 
     & = \frac{\sum_{i=1}^n g_i}{\zeta^k_t}
\end{aligned}
\end{equation}
where $\lim\limits_{\zeta^k_t \to \infty} \mathcal{A}(\hm{b}_{t-1}^k) = 0 $. Therefore, the bound of $\mathcal{A}(\hm{b}_{t-1}^k)$ is proven to be between $0$ and $1$.
\end{proof}

\clearpage
\section{Length Reward}\label{app: lr}
In the first subsection, we will introduce our step-wise length reward function and how it works. In the second part, we will discuss in details how it is designed.

\subsection{Our Step-wise Length Reward}
Our length reward function $\mathcal{R}(\zeta_t^k, Z)$ is to give a length score to a beam at each decoding step, so that its cumulative score $R(\zeta_T^k, Z)$ for a terminated beam could approach the maximum as its length $\zeta_T^k$ gets closer to the optimal length $Z$. Our designed length reward function is as follows:
\begin{equation}
    \mathcal{R}(\zeta_t^k, Z) = - \frac{\left|\zeta_t^k - \frac{Z}{\sqrt{2}}-0.5\right|}{Z}
    \label{eq:rat}
\end{equation}
Eq.~\ref{eq:joint} is thus expanded to:
\begin{equation}
\begin{aligned}
    J(\hm{b}_t,\hm{x}) 
     &= \log p(\hm{b}^k_{t-1}|\hm{x}) +\beta\left( \log A(\hm{b}^k_{t-1})+\gamma \sum_{m=1}^t\mathcal{R}(\zeta_m^k, Z)\right)+ \log p(v|\hm{x}, \hm{b}^k_{t-1})\\
     &=J(\hm{b}^k_{t-1},\hm{x})  +\log p(v|\hm{x}, \hm{b}^k_{t-1})+\beta \left( \log \mathcal{A}(\hm{b}^k_{t-1})+\gamma \mathcal{R}(\zeta_t^k, Z)\right),\ t \geq 2
     \label{eq:joint1}
\end{aligned}
\end{equation}
In this case, the final score of $\hm{y}^k$ is re-defined as
\begin{equation}
  \hm{y^*} = \argmax_{\hm{y}^k \in \mathcal{Y}}\ \frac{J(\hm{y}^k,\hm{x})}{|\hm{y}^k|-1}  
    \label{eq:final1}
\end{equation}
The original $R(\zeta_T^k, Z)$ in Eq.~\ref{eq:final} has been integrated to $J(\hm{y}^k,\hm{x})$, and exactly equals $\frac{\sum_{t=1}^T \mathcal{R}(\zeta_t^k, Z)}{|\hm{y}^k|-1}=\frac{\sum_{t=1}^{\zeta_T^k} \mathcal{R}(t, Z)}{\zeta_T^k}$ which gets the maximum at $\zeta_T^k=Z$. Related theorem and proof are presented in the next subsection.

\subsection{How It Is Designed}
\begin{figure}
\centering
\includegraphics[scale=0.7]{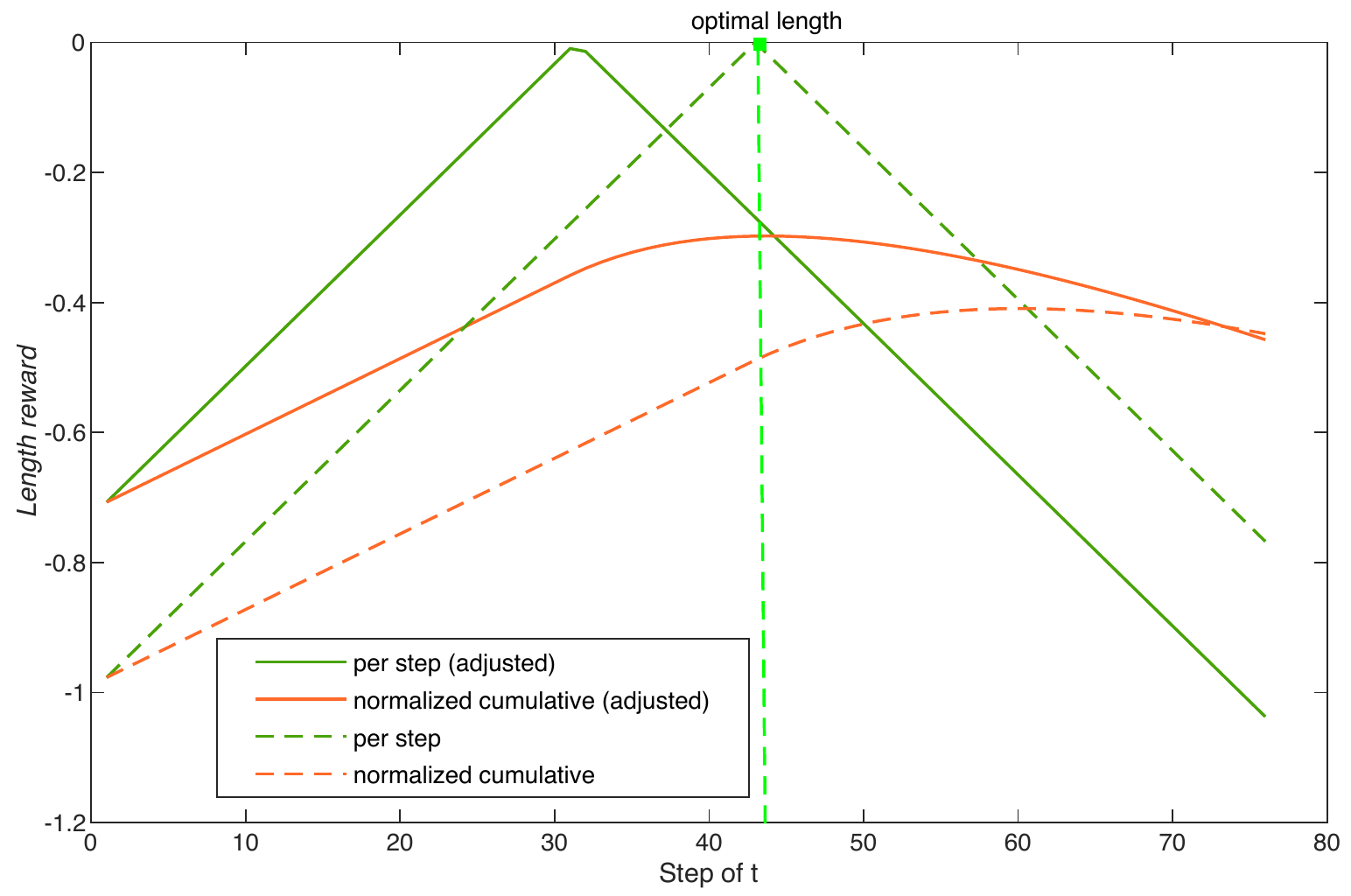}
\caption{Difference between length reward and adjusted length reward.}
\label{fig:plot2}
\end{figure}

 Intuitively, if we want to give a score to the beam length at each step, the score should reach the maximum when the beam length gets the optimality. Accordingly it can be defined as:
\begin{equation}
    \mathcal{R}(\zeta^k_t, Z) = - \frac{\left|\zeta_t^k - Z\right|}{Z},\ Z = \sum_{i=1}^n g_i
\end{equation}
Since $\zeta^k_t=t$, it is represented as:
\begin{equation}
    \mathcal{R}_t = - \frac{|t - Z|}{Z}
\end{equation}
Then, we test whether its cumulative value at the terminated step $R(\zeta^k_T, Z)$ reaches the maximum when $\zeta^k_T = Z$. The cumulative length reward is defined as
\begin{equation}
    R(\zeta^k_T, Z) = \frac{\sum_{t=1}^{\zeta^k_T} \mathcal{R}_t}{\zeta^k_T}
\end{equation}
If we replace $\zeta^k_T$ by $j$, it can be represented as:
\begin{equation}
    R_j = \frac{\sum_{t=1}^j \mathcal{R}_t}{j}
\end{equation}
 Unfortunately, when we plot the curve of $R_j$ as the orange dummy line in Figure~\ref{fig:plot2}, we find that  it reaches the maximum at a length longer than the optimal length. This prompts us to translate $\mathcal{R}_t$ from the green dummy line to the green full line, and the adjusted length reward is designed as:
\begin{equation}
    \mathcal{R}_t = - \frac{\left|t - \frac{Z}{\sqrt{2}}-0.5\right|}{Z}
    \label{eq:rat1}
\end{equation}
which is the same as Eq.~\ref{eq:rat}. After the adjustment, the normalized cumulative length reward $R_j$ (the orange solid line in Figure~\ref{fig:plot2}) peaks right at the optimal length.

\begin{customthm}{2}
Given $\mathcal{R}_t = - \frac{|t +D - Z|}{Z}$ ($t \geq 1$) and $R_j=\frac{\sum_{t=1}^j \mathcal{R}_t}{j}$ reaches the maximum when $j = Z$, then $D$ is approximate to $-\frac{Z}{\sqrt{2}} -0.5 + Z$.
\end{customthm}

\begin{proof}
Since $t \geq 1$ and $t \in \mathbb{Z}^+$, $\mathcal{R}_t$ can be regarded as two arithmetic sequences with difference of $\frac{1}{Z}$ and $-\frac{1}{Z}$, so that it is easy to calculate $\sum_{t=1}^j \mathcal{R}_{t}$. Then $R_j$ could be computed by:
\begin{equation}
    R_j = -\frac{1}{j}\left[\frac{(Z-D-1)(Z-D)}{2Z}  +\frac{(j+D-Z)(j+D-Z+1)}{2Z}\right]
\end{equation}
Hence the derivative of $R^a_j$ is:
\begin{equation}
    \frac{\partial R_j}{\partial j} = \frac{\frac{2(Z-D)^2-2(Z-D)}{j^2} - 1}{2Z}
\end{equation}
In addition, when $j=Z$, letting $ \frac{\partial R_j}{\partial j}=0$, we have:
\begin{equation}
    2D^2-(4Z-2)D + Z(Z-2) = 0
\end{equation}
If we solve this equation, we can get:
\begin{equation}
    D = \frac{4Z-2-\sqrt{8Z^2+4}}{4}
\end{equation}
Omitting the constant item in the radical due to $8Z^2 \gg 4$ under normal conditions, we can get:
\begin{equation}
    D = Z -0.5-\frac{Z}{\sqrt{2}}
\end{equation}
If we further take $D$ into $\mathcal{R}_t$, we can get Eq.~\ref{eq:rat1}.
\end{proof}
\clearpage

\section{Flow Chart: Global-aware vs. Beam Search}
\tikzstyle{startstop} = [rectangle,rounded corners, minimum width=3cm,minimum height=1cm,text centered, draw=black,text width =5cm]
\tikzstyle{io} = [trapezium, trapezium left angle = 70,trapezium right angle=110,minimum width=3cm,minimum height=1cm,text centered,draw=black,text width =3cm]
\tikzstyle{io1} = [trapezium, trapezium left angle = 70,trapezium right angle=110,minimum width=3cm,minimum height=1cm,text centered,draw=black,text width =2cm]
\tikzstyle{process} = [rectangle,minimum width=3cm,minimum height=1cm,text centered,text width =4cm,draw=black]
\tikzstyle{process1} = [rectangle,minimum width=3cm,minimum height=1cm,text centered,text width =4cm,draw=red]
\tikzstyle{decision} = [diamond,minimum width=3cm,minimum height=1cm,text centered,draw=black]
\tikzstyle{arrow} = [thick,->,>=stealth]
\tikzstyle{point} = [coordinate, on grid]

\begin{tikzpicture}[node distance=2cm]
\node (start) [startstop] {Enter global-aware inference};
\node (start1) [startstop, right of=start, node distance=8cm] {Enter beam search};
\node (input1) [io,below of=start] {Source and Predicted global attention distribution};
\node (input2) [io1,below of=start1] {Source document};
\node (process1) [process,below of=input1] { Unfinished sequences $\{\hm{b}^k\}^{\mathcal{K}}_{k=1}$ with scores $\{J^k\}_{k=1}^{\mathcal{K}}$};
\node (process1a) [process,below of=input2] { Unfinished sequences $\{\hm{b}^k\}^{\mathcal{K}}_{k=1}$ with scores $\{J^k\}_{k=1}^{\mathcal{K}}$};

\node (process2) [process1,below of=process1] {Calculate $\mathcal{A}(\cdot)$ for $\hm{b}^k$ by Eq.~\ref{eq:as}};
\node (process3) [process1,below of=process2] {Calculate $\mathcal{R}(\cdot)$ 
   for $\hm{b}^k$ by Eq.~\ref{eq:rat}};
\node (process4) [process,below of=process3] {Calculate the probability distribution of vocabulary $\mathcal{V}$ based on $\hm{b}^k$};
\node (process4a) [process,below of=process1a] {Calculate the probability distribution of vocabulary $\mathcal{V}$ based on $\hm{b}^k$};
\node (process5) [process,below of=process4] {Update sequences to$\{\hm{b}^i\}^{\mathcal{K} \times |\mathcal{V}|}_{i=1}$, and scores to $\{J^i\}_{i=1}^{\mathcal{K} \times |\mathcal{V}|}$ by Eq.~\ref{eq:joint1}};
\node (process6) [process,below of=process5] {Select $\mathcal{K}$ unfinished sequences with highest scores};
\node (process5a) [process,below of=process4a] {Update sequences to$\{\hm{b}^i\}^{\mathcal{K} \times |\mathcal{V}|}_{i=1}$, and scores to $\{J^i\}_{i=1}^{\mathcal{K} \times |\mathcal{V}|}$ by Eq.~\ref{eq: accumu}};
\node (process6a) [process,below of=process5a] {Select $\mathcal{K}$ unfinished sequences with highest scores};
\node (decision1) [decision,below of=process6,yshift=-0.5cm] {Finished};
\node (point) [point, right of=decision1, node distance=3cm]{--};
\node (stop) [startstop,below of=decision1,yshift=-0.5cm] {Select the final hypothesis with the highest score by Eq.~\ref{eq:final1}};
\node (decision2) [decision,below of=process6a,yshift=-0.5cm] {Finished};
\node (point1) [point, right of=decision2, node distance=3cm]{--};
\node (stop1) [startstop,below of=decision2,yshift=-0.5cm] {Select the final hypothesis with the highest score by Eq.~\ref{eq: beam-search}};

\draw [arrow] (start) -- (input1);
\draw [arrow] (start1) -- (input2);
\draw [arrow] (input1) -- (process1);
\draw [arrow] (input2) -- (process1a);
\draw [arrow] (process1) -- (process2);
\draw [arrow] (process2) -- (process3);
\draw [arrow] (process3) -- (process4);
\draw [arrow] (process1a) -- (process4a);
\draw [arrow] (process4) -- (process5);
\draw [arrow] (process4a) -- (process5a);
\draw [arrow] (process5) -- (process6);
\draw [arrow] (process5a) -- (process6a);
\draw [arrow] (process6) -- (decision1);
\draw [-] (decision1) --  node[above] {No}(point);
\draw [arrow] (point) |-  (process1);
\draw [arrow] (decision1) -- node[left] {Yes} (stop);
\draw [arrow] (process6a) -- (decision2);
\draw [-] (decision2) --  node[above] {No}(point1);
\draw [arrow] (point1) |-  (process1a);
\draw [arrow] (decision2) -- node[left] {Yes} (stop1);

\end{tikzpicture}

\clearpage
\section{Scoring Mechanism Comparison}
\label{app: smc}
\begin{table}[htbp]\scriptsize
    \centering
     \caption{Scoring Mechanism Comparison. Training: whether the score function is a part of loss function of the seq-to-seq model. Inference:  whether the score function is a part of the hypothesis score. Desired Situation: the score function reaches the optimum in this situation. $\dagger$: the function described in the paper \cite{gehrmann-etal-2018-bottom} is only activated at the termination, but their code added a step-wise modification on it.}
    \begin{tabular}{l|c|c|c|c|c}
    \toprule
        {\bf Article} & {\bf Score Function} & {\bf Training} & {\bf Inference} & {\bf Step-wise} & {\bf Desired Situation}  \\\midrule
        \cite{Wu2016Google} &  $\sum_{i=1}^n \log \min\left(\sum_{t=1}^T\alpha_{t,i}, 1\right)$&\XSolid&\Checkmark&\XSolid& $\sum_{t=1}^T\alpha_{t,i} \geq 1$\\\midrule
        \cite{li2018simple} & $\sum_{i=1}^n \log \min\left(\sum_{t=1}^T\alpha_{t,i}, \beta\right)$& \XSolid&\Checkmark&\XSolid& $\sum_{t=1}^T\alpha_{t,i} \geq \beta$\\\midrule
        \cite{See2017} & $-\sum_{i=1}^n \min \left(\alpha_{t,i}, \sum_{m=1}^{t-1}\alpha_{m,i}\right)$&\Checkmark&\XSolid&\Checkmark& $\sum_{m=1}^{t-1}\alpha_{m,i} <\alpha_{t,i}$ \\\midrule
        \cite{li-etal-2018-improving-neural} & $1-\sum_{i=1}^n \min \left(\alpha_{t,i}, \sum_{m=1}^{t-1}\alpha_{m,i}\right)$&\Checkmark&\Checkmark&\Checkmark&$\sum_{m=1}^{t-1}\alpha_{m,i} <\alpha_{t,i}$ \\\midrule
        \cite{gehrmann-etal-2018-bottom} & $n-\sum_{i=1}^n \max \left(\sum_{m=1}^t \alpha_{t,i}, 1\right)$&\XSolid&\Checkmark&\Checkmark$\dagger$&$\sum_{m=1}^t \alpha_{m,i} \leq 1$\\\midrule
        Ours & $\frac{\sum_{i=1}^n \min\left(\sum_{m=1}^t\alpha_{m,i}, g_i\right)}{\sum_{i=1}^{n}\sum_{m=1}^t\alpha_{m,i}} + \gamma \mathcal{R}(\zeta_t^k, Z)$&\XSolid&\Checkmark&\Checkmark& \tabincell{c}{$\sum_{m=1}^t\alpha_{m,i} < g_i, t <T$ \\ \textbf{and} $\sum_{m=1}^T\alpha_{m,i} = g_i$}\\
        \bottomrule
    \end{tabular}
    
    \label{tab:penalty com}
\end{table}

We summarize the characteristics of each score function in Table \ref{tab:penalty com}, where $\alpha_{t,i}$ refers to the cross attention that the $t_{th}$ generated token pays to the $i_{th}$ source token. Previous works \cite{Wu2016Google, li2018simple,See2017,li-etal-2018-improving-neural,gehrmann-etal-2018-bottom} design the score function based on some assumptions which may be invalid. Specifically, \cite{Wu2016Google}, \cite{li2018simple} and \cite{gehrmann-etal-2018-bottom} use the same constant (1 or $\beta$) to constrain all source tokens, based on the assumption that the minimum/maximum attention allocated to each token is the same. \cite{See2017} and \cite{li-etal-2018-improving-neural} assume the attention assigned before $t$ should be lower than the attention of $t$. Obviously, both assumptions may often not hold in practice. By contrast, our global scoring mechanism does not depend on any assumption but follows the nature of generation, i.e., the growing local attention should not surpass the global attention during inference and exactly reach it at the termination.

\clearpage
\section{Discussions}
\subsection{Predictability of Global Attention}\label{app: dp}
\begin{figure}[htbp]
\centering
\begin{minipage}[t]{0.44\textwidth}
\includegraphics[width=5.8cm]{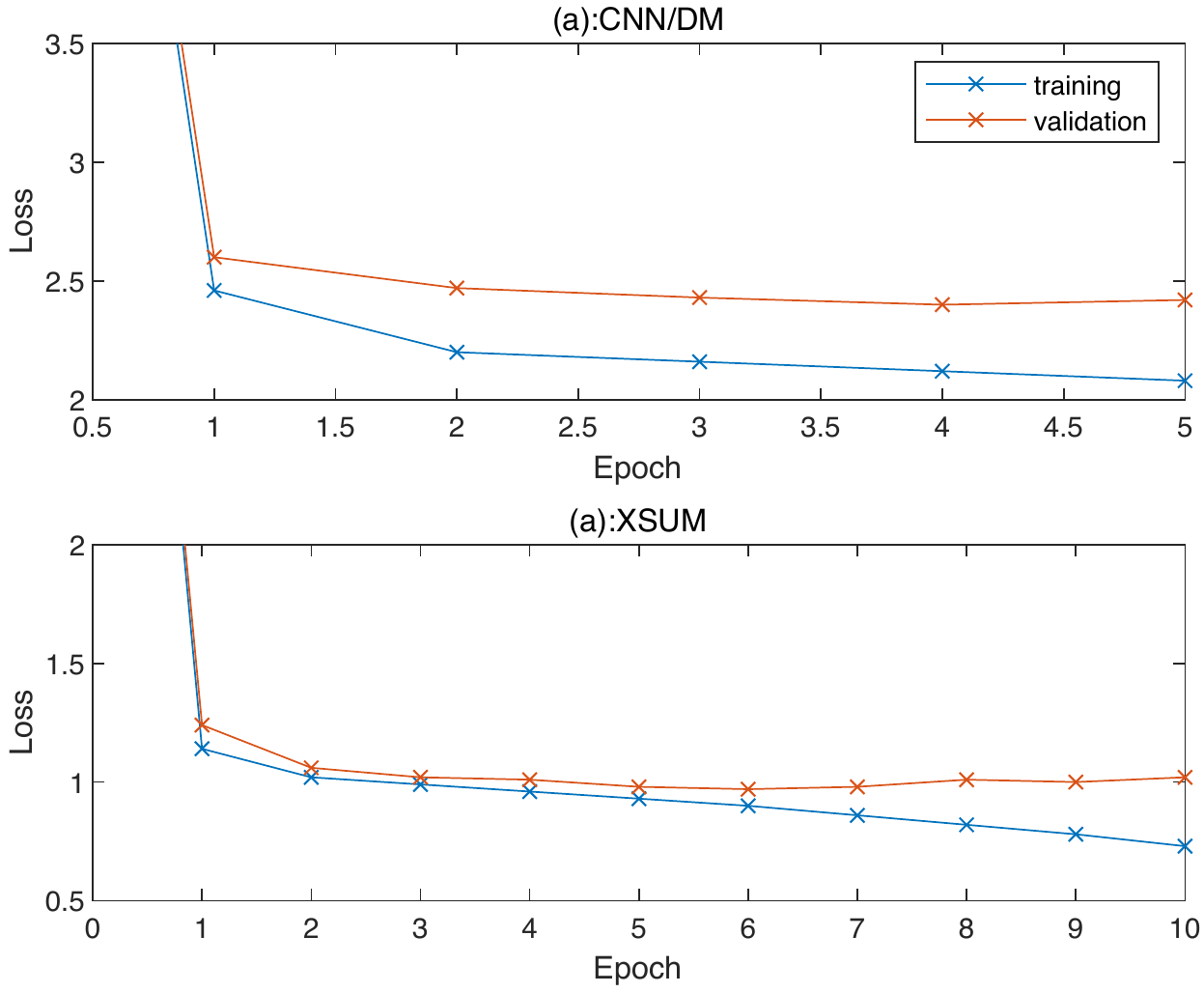}
\caption{Loss trend.}
\label{fig:loss trend}
\end{minipage}
\hspace{0.3in}
\begin{minipage}[t]{0.44\textwidth}
\includegraphics[width=5.8cm]{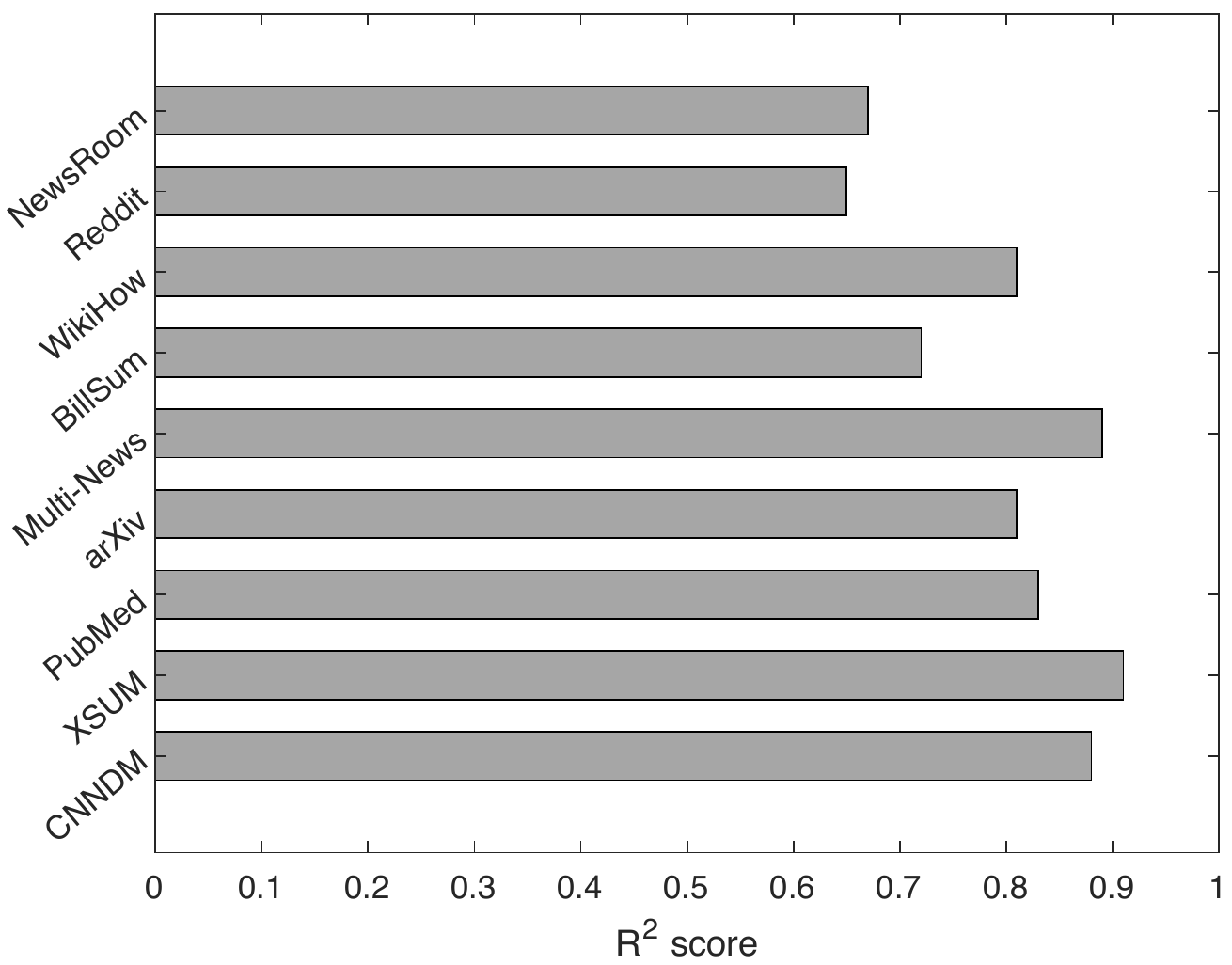}
\caption{Fitting of att-prediction model.}
\label{fig:r2}
\end{minipage}
\end{figure}

We plot the decline trend of training loss and validation loss on CNN/DM and XSUM to Figure \ref{fig:loss trend}. It is observed that training loss has been falling significantly, while validation loss has an upward trend at later epochs.

On the other hand, we randomly select $1,000$ test samples from each dataset and plot their average $R^2$ scores of the predicted optimal attention distributions in Figure~\ref{fig:r2}. For each sample, its $R^2$ score is equal to $1 - \frac{\sum_{i=1}^{n} (o_i - \hat{o}_i)^2}{\sum_{i=1}^{n} (o_i - \bar{o})^2}$, where $R^2 \in [-\infty,1]$. A higher $R^2$ signifies better model fitting, while $R^2 < 0$ means the model is entirely invalid. 

\subsection{Degradation of Beam Search}
\begin{figure}[htbp]
\centering
\includegraphics[scale=0.65]{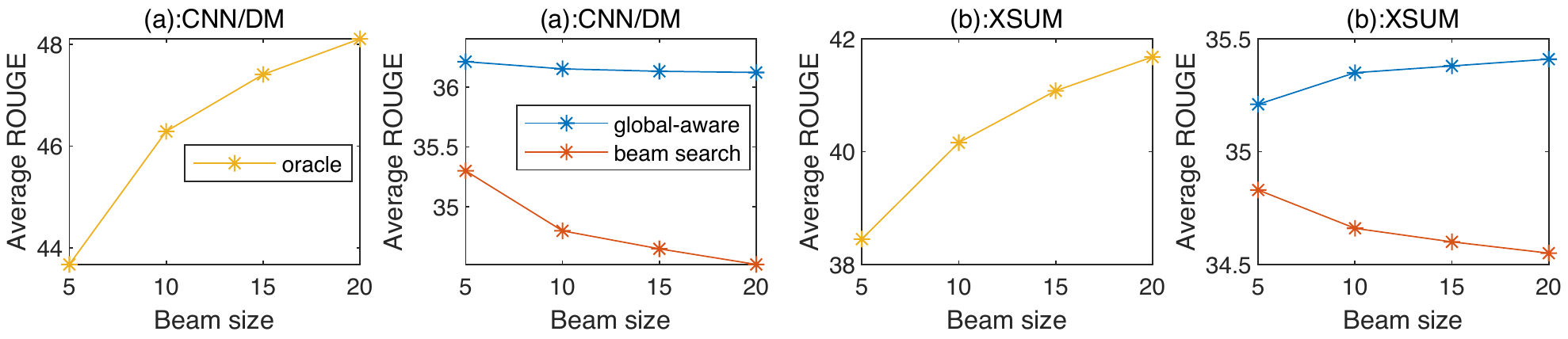}
\caption{Degradation of beam search.}
\label{fig:enlarge}
\end{figure}

It is widely known that text quality generally degrades with increased beam size~\cite{cohen2019empirical,ott2018analyzing,meister2020if}. The degradation originates from the gap between training and inference, where teacher forcing training only need  ensure the local optimality, but beam search pursues the global optimality~\cite{wiseman-rush-2016-sequence, ranzato2016sequence, zhang-etal-2019-bridging}. The proposed algorithm reduces the gap by informing beam search how the global optimal hypothesis attends to the source so that beam search can be calibrated to pursue a more reasonable global optimum instead of that obtained by teacher forcing training. We enlarge the beam size and find that the global attention distribution can guide beam search to overcome degradation, though the practical improvement might not be significant due to the prediction bias.
According to Figure \ref{fig:enlarge}, the average ROUGE $F_1$ score of beam search summaries declines sharply as beam size increases. Even if there is a prediction bias, the global-aware inference manages to resist degradation effectively with increased beam sizes. Moreover, a significant boosting in ROUGE scores is observed as beam size increases in the ORACLE global-aware inference, where the true potential of the algorithm is reflected.

\subsection{Length Comparison}
\begin{table*}[htbp]\scriptsize
  \centering
  \caption{The average summary length. $\dagger$: consisting of shorter beam search and global-aware$\dagger$. *: without length reward. In addition to the two datasets, the numbers closer to the reference length are bolded while shorter lengths are underlined. All reference summaries are truncated to 256 tokens.}
  \vspace{0.05in}
    \begin{tabular}{lcccccccc|cc}
    \toprule
    {\bf Token Length} & {\bf CNNDM}   & {\bf Bill} & {\bf Multi} & {\bf Wiki} & {\bf Reddit}& {\bf NRoom} & {\bf Pub} & {\bf arXiv}& {\bf CNNDM}$\dagger$ & {\bf XSUM}*\\
    \midrule
    {\it reference} & 67.5  & 176.5 & 232.9 & 70.5 & 26.2 & 34.4 & 214.7 & 191.3 & 67.5 & 26.1\\
    \midrule
    {\it beam search} & 82.3  & \uline{162.5} & \uline{219.4} & {\bf 49.8} & 36.5 & 46.7 & {\bf 205.8} & {\bf 176.1}& 78.2 & 21.8 \\
    {\it global-aware} & \uline{{\bf 64.9}}  & {\bf 165.6} & {\bf 223.7} & \uline{46.2} & \uline{\bf 20.8} & \uline{\bf 31.2} & \uline{200.4} & \uline{166.4} & 53.2 & 22.2\\
    \bottomrule
    \end{tabular}
     \label{tab:len}
\end{table*}
As shown in Table~\ref{tab:len}, even with empirical hyper-parameters,  our method  often leads to more concise summary with its length closer to the reference length on many datasets. By contrast, beam search, which conducts searching for length penalty and length constraints, generates many lengthy summaries on some datasets. For example, In CNN/DM, Reddit, and NewsRoom datasets, beam search must extend summaries to obtain higher ROUGE scores, making the generated text more bloated. Overall, in most cases, the global-aware inference could produce summaries of higher quality in a more concise way. 

\subsection{Are the Outputs Different from Beam Search from the Beginning?}\label{app: dd}
\begin{figure}
\centering
\includegraphics[scale=0.7]{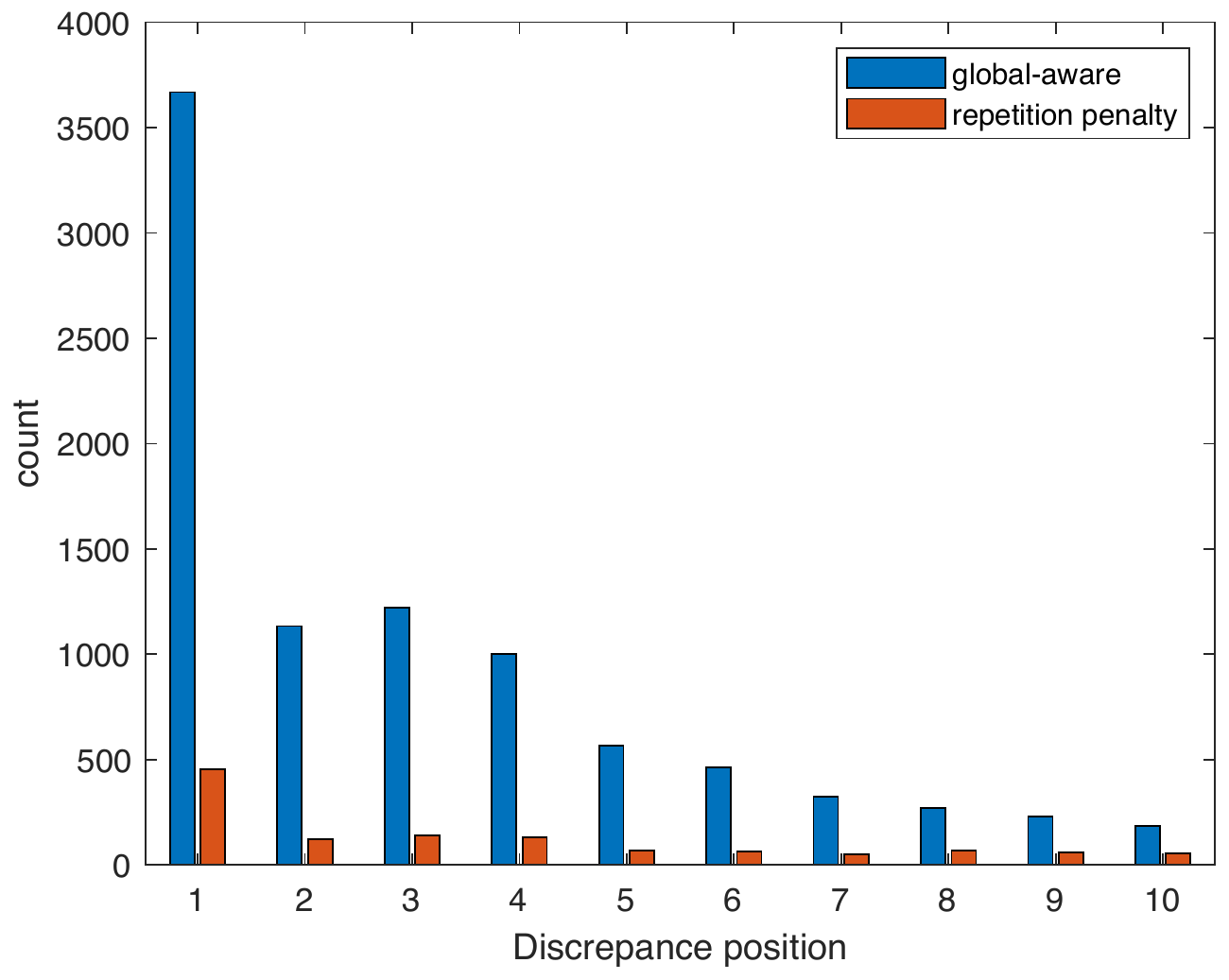}
\caption{ The position that the summary starts to differ from the beam search summary.}
\label{fig:dis}
\end{figure}

 One may question whether these hypotheses are just the subsets or extensions of beam search hypotheses. The answer lies in the selection of $\beta$. Specifically, the global-aware with a greater $\beta$ (such as $\beta=12$ for CNN/DM) generates quite different summaries. We plot the count distribution of discrepancy positions between beam search and global-aware inference in Figure~\ref{fig:dis}. It is interesting to find that about one-third of CNN/DM test samples (total count: $11,490$) start to differ at the first word, and most start to differ at the first $5$ words. In other words, the global-aware inference generates texts in a way different from beam search that leads to higher scores.

\subsection{Newly Generated Words}

Table~\ref{tab:novel} shows that summaries generated by our method appear more creative. 
Generally, a greater proportion of novel words implies that the generated text resembles more human-writing, at the same time increases the risk of deviating from reference. 

\subsection{Inference Speed}
As presented in Table \ref{tab:speed}, the speed of the global-aware inference is comparable with beam search.

\begin{table}[htbp]
\begin{minipage}[t]{0.45\textwidth}
    \caption{The percentage of words in the summary but not in the source.}
  \centering
    \begin{tabular}{lc}
    \toprule
    {\bf CNN/DM} & {\bf \%Novel}\\
    \midrule
    {\it reference} & 14.8\\
    {\it beam search} & 4.5\\
    {\it global-aware } & {\bf 6.7}\\
      \bottomrule
    \end{tabular}
     \label{tab:novel}
\end{minipage}
\hspace{0.3in}
 \begin{minipage}[t]{0.45\textwidth}
     \centering
      \caption{Inference speed}
      \vspace{0.17in}
    \begin{tabular}{lcc}
    \toprule
    {\bf CNN/DM}& {\bf Tokens/s}\\
    \midrule
    {\it beam search}  & {\bf 25.6}\\
    {\it global-aware}  & 23.2\\
    {\it repetition penalty} & 19.1\\
   
      \bottomrule
    \end{tabular}
     \label{tab:speed}
 \end{minipage}
\end{table}

\subsection{Our Advantages over Predicting Future Rewards}
A previous study \cite{DBLP:journals/corr/LiMJ17} guides inference by predicting the future automatic matrix score at each decoding step. There are two main disadvantages of training such predictor, (a) there are too many training samples especially when the target sequence is long, making it hard to train; (b) this method forces us to predict every decoding step, greatly slowing the inference speed.
By contrast, one appealing benefit of our approach is its high efficiency, since our method only need predict once but can calibrate beam search at each step, which improves beam search from scratch with very little burden of training and inference. 

\subsection{Can 
Our Method Work Together with Attention Head Masking?}
There is no doubt that the answer is yes because the attention head masking \cite{cao2021attention} still need beam search to generate summaries, and our approach is an alternative to beam search. However, we do not recommend the joint usage of the two due to the following three reasons. First, the roles of the two overlap; second, the loss of context information due to masking will negatively affect the global-aware inference; third, adding attention head masking will impose additional burden on training and hyper-parameter searching.

\subsection{Can Predicted Attention Distribution still Lead to the Same Theoretical Result?}
Through explaining that the global attention and the predicted attention have the same value range, we will simply prove that the predicted attention distribution is also applicable to the theory in the paper. Since each attention value in the global attention distribution is a cumulative value of the attention probability $\in [0, 1]$ given by the reference tokens, the lower bound tends to be $0$ and the upper bound is $1$ $*$ the number of reference tokens.  When the reference is infinite, the upper bound is $\infty$. Therefore, we only need  ensure that the predicted attention value is larger than $0$ which is constrained by $\exp$.

\section{Case Study}
Examples are presented in the following pages. We sample some good cases and some bad cases of global-aware inference by the ROUGE-1 $F_1$ score.
\clearpage

\begin{table}[htbp]\scriptsize
    \centering
     \caption{Generated summaries on CNN/DM sampled by ROUGE-1 $F_1$ score.}
    \begin{tabular}{p{43pt}|p{330pt}}
    \toprule
        {\it Reference} & The ramp agent fell asleep in the plane's cargo hold . He can no longer work on Alaska Airlines flights .\\
        \midrule
      {\it Beam search}  (R-1: 25.05)&"I\'m inside a plane and I feel like it\'s up moving in the air," the caller says . "There could be a person in there so we\'re going to come back around," the pilot tells air traffic control . The ramp agent is an employee of Menzies Aviation, a contractor for Alaska Airlines . The airline says the man has been permanently banned from working on planes .
        \\
      \midrule
      {\it Global-aware}  (R-1: 43.29) & A ramp agent fell asleep in the cargo hold of an Alaska Airlines flight . The crew and passengers reported unusual banging from the belly of the Boeing 737 . The pilot radioed air traffic control and said he would make an emergency landing .\\
       \midrule
      {\it Global-aware$\dagger$} (R-1: 50.81)&A ramp agent falls asleep in the cargo hold of an Alaska Airlines plane . The plane makes an emergency landing in Seattle . The man has been banned from working on Alaska Airlines planes .
        \\
        \toprule
         {\it Reference} & Three people killed; five wounded in attack on attorney general's office in Balkh province . Staff and civilians have been rescued as gunmen engaged Afghan security forces .
\\
        \midrule
      {\it Beam search}  (R-1: 38.78)& A group of armed assailants stormed into the attorney general's office in Balkh province, northern Afghanistan . Two police officers and a security guard of the provincial attorney general's office were among the dead . Most staff members and civilians have been rescued, but an exchange of fire between Afghan security forces and the assailants is ongoing .
        \\
      \midrule
      {\it Global-aware}  (R-1: 54.00) & Three people were killed and five others were wounded in an attack on the attorney general's office . An exchange of fire between security forces and the assailants is ongoing ."\\
       \midrule
      {\it Global-aware$\dagger$} (R-1: 44.00)&Gunmen storm attorney general's office in northern Afghanistan . Three people killed, five others wounded .
        \\
          \toprule
         {\it Reference} & Six young Minnesotans conspired to sneak into Syria and join ISIS "by any means necessary," prosecutors say . The men, ages 19 to 21, were arrested Sunday . They plotted for 10 months, the U.S. attorney for the District of Minnesota says .
\\
        \midrule
      {\it Beam search}  (R-1: 47.68)&The group of friends, ages 19 to 21, were arrested Sunday . They plotted for 10 months, U .S . Attorney Andrew M . Luger says . Recruiting for the ISIS terrorist network is a problem in Minnesota's Somali community . Four of the men who were arrested appeared in federal court on Monday but did not enter a plea .
        \\
      \midrule
      {\it Global-aware}  (R-1: 61.97) & Six men, ages 19 to 21, were arrested Sunday . They plotted for 10 months, U .S . attorney says . Recruiting for the ISIS terrorist network is a problem in Minnesota's Somali community .
\\
       \midrule
      {\it Global-aware$\dagger$} (R-1: 38.92)&The six men were arrested Sunday . They are accused of plotting to join ISIS in Syria . Recruiting for ISIS is a problem in Minnesota's Somali community .
        \\
        \toprule
         {\it Reference} & Raul Reyes: In seeking Latino vote, Marco Rubio his own worst enemy on two key issues: immigration reform, Cuba relations . He says on health care, climate change and other issues, he breaks from Latinos' positions. Polls show they don't favor him .
\\
        \midrule
      {\it Beam search}  (R-1: 35.71)&Ruben Navarrette: Marco Rubio has been his own worst enemy on immigration and Cuba . Navarrette: Rubio has embraced a typical conservative approach to immigration . He says Rubio's political philosophy will be a tough sell to Hispanics . Navarrette: Unless he can offer new ideas, his climb to the Republinomination will be steep .
        \\
      \midrule
      {\it Global-aware}  (R-1: 36.50) & Ruben Navarrette: Marco Rubio is running for the GOP presidential nomination . Navarrette: Rubio has been his own worst enemy on immigration reform and Cuba relations . He says Rubio has embraced a typical conservative approach to immigration . Navarrette: Rubio's political philosophy will be a tough sell to Hispanics .
\\
       \midrule
      {\it Global-aware$\dagger$} (R-1: 40.11)&Ruben Navarrette: Marco Rubio is running for the GOP presidential nomination . Navarrette: Rubio has been his own worst enemy on immigration reform and Cuba relations . He says Rubio has embraced a typical conservative approach to immigration .
        \\
         \toprule
         {\it Reference} & Police: Yuhei Takashima, 64, says he had sex with girls as young as 14 in Philippines . Officers seize nearly 150,000 photos that the former principal kept of his activities .
\\
        \midrule
      {\it Beam search}  (R-1: 52.48)&Yuhei Takashima, 64, says he paid for sex with more than 12,000 women in the Philippines . Police seized 147,600 photos that Takashima took of his activities over the years . The ages of the prostitutes he hired ranged from 14 to over 70, he says .
        \\
      \midrule
      {\it Global-aware}  (R-1: 37.33) & Police say Yuhei Takashima, 64, paid for sex with more than 12,000 women . He took 147,600 photos of his activities over more than a quarter of a century, police say .
\\
       \midrule
      {\it Global-aware$\dagger$} (R-1: 49.28)&Police: Yuhei Takashima, 64, paid for sex with more than 12,000 women in the Philippines . He took 147,600 photos of his activities, police say .
        \\
         \toprule
         {\it Reference} & Kabul faces uncertain future as NATO presence -- and the money that came with it -- fades away . Interpreters are out of work, NATO trucks sit idle on roads, restaurants are empty .
\\
        \midrule
      {\it Beam search}  (R-1: 21.95)&Kabul is a city swollen in size but shrunken in scope, anxiously awaiting what comes next . Vast supply chains once kept 120,000 troops fed and watered . Now the bases are gone, and the trucks that once supplied millions sit still . One road, forever pot-holed in the past decade, is now being covered over by Afghans .
        \\
      \midrule
      {\it Global-aware}  (R-1: 16.12) & Afghanistan's capital, Kabul, has been transformed since NATO arrived in 2001 . The city's population has swelled to five times what it was when the U .S . arrived . But the city has also shrunk in scope, with many waiting to see what comes next .
\\
       \midrule
      {\it Global-aware$\dagger$} (R-1: 20.39)&"Kabul, the capital of Afghanistan, has been transformed by the U .S . troop withdrawal . Roads are lined with the detritus of America's war here . Vast supply chains once kept 120,000 troops fed and watered .
        \\
        \bottomrule
    \end{tabular}
    
\end{table}

\begin{table}[htbp]\scriptsize
    \centering
     \caption{Generated summaries on BillSum sampled by ROUGE-1 $F_1$ score.}
    \begin{tabular}{p{43pt}|p{330pt}}
    \toprule
        {\it Reference} & Directs the Secretary of Health and Human Services to establish a process under which a physician may request, in writing from a carrier, assistance in addressing questionable codes and procedures under the medicare program. Sets forth provisions concerning: (1) policy development regarding evaluation and management guidelines; and (2) medicare overpayments.\\
        \midrule
      {\it Beam search}  (R-1: 27.61)&Prohibits the Health Care Financing Administration (HCFA) from implementing any new evaluation and management guidelines under the Medicare program unless HCFA: (1) has provided for an assessment of the proposed guidelines by physicians; (2) has established a plan for improving participation of physicians; (3) has carried out a minimum of four pilot projects in at least four different HCFA regions to test such guidelines; and (4) finds that the objectives will be met in the implementation of such guidelines . Requires each pilot project to: (1) be of sufficient length to allow for preparatory physician and carrier education, analysis, and use and assessment of potential E\&M guidelines; and (2) be conducted, throughout the planning and operational stages of the project, in consultation with national and State medical societies .
        \\
      \midrule
      {\it Global-aware}  (R-1: 40.00) & Directs the Secretary of Health and Human Services to establish a process under which a physician may request, in writing from a carrier, assistance in addressing questionable codes and procedures under the Medicare program . Requires the carrier to respond in writing within 30 business days and allows a written statement to be used as proof against a future audit or overpayment under the Medicare program . Requires the Administrator of the Health Care Financing Administration to restore the toll-free telephone hotline so that physicians may call for information and questions about the Medicare program .Prohibits the Secretary from implementing any new evaluation and management guidelines under the Medicare program, unless the Health Care Financing Administration: (1) has provided for an assessment of the proposed guidelines by physicians; (2) has established a plan for improving participation of physicians; (3) \dots\\
        \toprule
         {\it Reference} & Reduce Expenditures in Nuclear Infrastructure Now Act or the REIN-IN Act - Prohibits the obligation or expenditure of funds authorized to be appropriated to the Department of Defense (DOD) for FY2014-FY2023: (1) for the research, development, test, and evaluation (RDT\&amp;E) or procurement of a long-range penetrating bomber aircraft; (2) to procure an SSBN-X submarine (and prohibits the use of such funds for FY2024 and thereafter to procure more than eight such submarines); or (3) for the RDT\&amp;E or procurement of a new intercontinental ballistic missile (ICBM). Prohibits the obligation or expenditure of funds authorized to be appropriated for FY2014 or thereafter for DOD or the Department of Energy (DOE): (1) to make the F-35 Joint Strike Fighter aircraft capable of carrying nuclear weapons; (2) until the Secretary of Defense and the Secretary of Energy jointly certify that the total cost of the B61 life extension program has been reduced to not more than \$5 billion; (3) for the W78 life extension program; (4) for the mixed oxide fuel fabrication facility project; (5) to replace the chemistry and metallurgy research building at Los Alamos National Laboratory, Los Alamos, New Mexico; or (6) for the uranium processing facility at the Y-12 National Security Complex, Oak Ridge, Tennessee. Prohibits Navy forces, beginning in FY2020, from including more than eight operational ballistic-missile submarines available for deployment . Prohibits the \dots 
\\
        \midrule
      {\it Beam search}  (R-1: 56.48)&Reduce Expenditures in Nuclear Infrastructure Now Act or the REIN-IN Act Prohibits using funds appropriated to the Department of Defense (DOD) for FY2014-FY2023: (1) for the research, development, test, and evaluation or procurement of a long-range penetrating bomber aircraft; (2) to make the F-35 Joint Strike Fighter aircraft capable of carrying nuclear weapons; (3) until the Secretary of Defense and the Secretary of Energy jointly certify that the total cost of the B61 life extension program has been reduced to not more than \$5 billion; (4) for the W78 life extension program; (5) for the reduction of nuclear-armed submarines, beginning in FY2020; or (6) for the National Nuclear Security Administration for FY2024 .
        \\
      \midrule
      {\it Global-aware}  (R-1: 67.03) &Reduce Expenditures in Nuclear Infrastructure Now Act or the REIN-IN Act Prohibits the obligation or expenditure of funds authorized to be appropriated to the Department of Defense (DOD) for FY2014-FY2023: (1) for the research, development, test, and evaluation or procurement of a long-range penetrating bomber aircraft; (2) to make the F-35 Joint Strike Fighter aircraft capable of carrying nuclear weapons; (3) until the Secretary of Defense and the Secretary of Energy jointly certify that the total cost of the B61 life extension program has been reduced to not more than \$5 billion; (4) for the W78 life extension program; (5) for the reduction of Nuclear-Armed Submarines, beginning in FY2020; (6) for the SSBN-X submarines; or (7) for the mixed oxide fuel fabrication facility project . Prohibits the obligation or expenditure of funds authorized to be appropriated for FY2014-FY2023 for DOD: (1) to maintain more than 200 intercontinental ballistic missiles (ICBMs), (2) to maintain more than 250 submarine-launched ballistic missiles, (3) for the research, development, test, and evaluation or procurement of a new ICBM, or (4) for the uranium processing facility at the Y-12 National Security Complex, Oak Ridge, Tennessee . Prohibits the \dots 
\\
 \toprule
         {\it Reference} & Medicare Part D Drug Class Protection Act of 2007 - Amends part D (Voluntary Prescription Drug Benefit Program) of title XVIII (Medicare) of the Social Security Act to require that Medicare prescription drug plans using formularies cover all drugs included in six specified therapeutic categories. Sets forth special requirements for reconsideration of coverage determinations, and appeals for drugs included in such categories. Establishes reporting requirements for drugs in these categories.
\\
        \midrule
      {\it Beam search}  (R-1: 73.59)& Amends part D (Voluntary Prescription Drug Benefit Program) of title XVIII (Medicare) of the Social Security Act to require prescription drug formularies to cover all drugs in six specified therapeutic categories and classes .
        \\
      \midrule
      {\it Global-aware}  (R-1: 68.45) &Amends part D (Voluntary Prescription Drug Benefit Program) of title XVIII (Medicare) of the Social Security Act to require that the prescription drug formulary include, subject to specified requirements, all or substantially all drugs in each of six specified therapeutic categories of covered Medicare part D drugs: (1) Immunosupessants; (2) Antidepressants; (3) Anticonvulsants; and (4) Antiretrovias . Provides for special coverage of drugs included in specified therapeutic categories during determinations, reconsiderations, and appeals .\\
      \bottomrule
     \end{tabular}
    
\end{table}

\begin{table}[htbp]\scriptsize
    \centering
     \caption{Generated summaries on arXiv sampled by ROUGE-1 $F_1$ score.}
    \begin{tabular}{p{43pt}|p{330pt}}
    \toprule
        {\it Reference} &  in this paper , we develop a structure - preserving discretization of the lagrangian framework for electromagnetism , combining techniques from variational integrators  and  discrete differential forms . this leads to a general family of variational , multisymplectic numerical methods for solving maxwell s equations that automatically preserve key symmetries and invariants  .    in doing so   , we demonstrate several new results , which apply both to some well - established numerical methods and to new methods introduced here  .   first , we show that yee s finite - difference time - domain ( fdtd ) scheme , along with a number of related methods , are multisymplectic and derive from a discrete lagrangian variational principle  .   second , we generalize the yee scheme to unstructured meshes , not just in space but in @xmath0-dimensional spacetime  .   this relaxes the need to take uniform time steps , or even to have a preferred time coordinate at all  . finally , as an example of the type of methods that can be developed within this general framework , we introduce a new asynchronous variational integrator ( avi ) for solving maxwell s equations  .   these results are illustrated with some prototype simulations that show excellent energy and conservation behavior and lack of spurious modes , even for an irregular mesh with asynchronous time stepping  .
\\
        \midrule
      {\it Beam search}  (R-1: 37.82)&the yee scheme is one of the most successful numerical methods used in the field of computational electromagnetics , particularly in the area of microwave problems  .   although it is not a  high - order '' method , it is still preferred for many applications because it preserves important structural features of maxwell s equations that other methods fail to capture  . in this paper   , we show that these desirable properties are direct consequences of the variational and discrete differential structure of the yee scheme , which mirrors the geometry of maxwell s equations  .   moreover , we show how to construct other variational methods that , as a result , share these same numerical properties , while at the same time applying to more general domains  .
        \\
      \midrule
      {\it Global-aware}  (R-1: 45.96) & the yee ( finite - difference time - domain ) scheme is one of the most successful numerical methods used in the field of computational electromagnetics , particularly in the area of microwave problems  . in this paper   , we show that these desirable properties are direct consequences of the variational and discrete differential structure of the yee scheme , which mirrors the geometry of maxwell s equations  .   moreover , we show how to construct other variational methods that , as a result , share these same numerical properties , while at the same time applying to more general domains  .   the key symmetry of maxwell s equations is not rotational or translational symmetry , as in mechanics , but a differential gauge symmetry . without taking additional care to preserve this gauge structure , even variational integrators can not be expected to capture the geometry of maxwell s equations  .   this differential / gauge structure also turns out to be important for the numerical performance of the method , and is one of the hallmarks of the yee scheme  .   this differential / gauge structure also turns out to be important for the numerical performance of the method , and is one of the hallmarks of the ye \dots \\
        \toprule
         {\it Reference} & we present a new general mechanism to generate curvature perturbations after the end of the slow - roll phase of inflation  .   our model is based on the simple assumption that the potential driving inflation is characterized by an underlying global symmetry which is slightly broken  . 
\\
        \midrule
      {\it Beam search}  (R-1: 36.88)&most inflationary models assume that there is only one scalar field involved in the dynamics of inflation  . in this work   we point out a qualitatively new effect that might arise if one relaxes the assumption of a single dynamical field  . in a multi - field scenario in which the inflationary potential is characterized by a broken symmetry , the quantum fluctuations generated during the inflationary stage represent fluctuations in the initial conditions for the dynamics of the inflaton in the subsequent stage , thus implying that the background dynamics after the slow - roll phase has ended will differ in different regions of the universe  .   since the background fields are coupled to the other fields into which they decay , the fluctuations generated during the slow - roll phase will affect the subsequent decay process  .   the present work , assuming that the inflaton decay into other fields through the non - perturbative process of preheating , is then aimed to understand whether isocurvature inflaton fluctuations , generated during the slow - roll stage , can lead to perturbations of the background metric through variations of the preheating efficiency  .   while the generation of curvature perturbations during the stages following the slow - roll phase has already been considered in some works , the \dots
        \\
      \midrule
      {\it Global-aware}  (R-1: 53.46) &we show that in a multi - field scenario in which the inflationary potential is characterized by a broken symmetry , quantum fluctuations generated during the slow - roll stage of inflation may affect the efficiency of the subsequent preheating process  .   we obtain a general formula for the curvature perturbations generated from an inhomogeneous preheating efficiency related to the quantum fluctuations produced during inflation  .   the present work is then applied to the case of a broken @xmath0 symmetry  .
\\
 \toprule
         {\it Reference} & in this work we investigate an influence of decoherence effects on quantum states generated as a result of the cross - kerr nonlinear interaction between two modes  . for markovian losses ( both photon loss and dephasing ) ,   a region of parameters when losses still do not lead to destruction of non - classicality is identified  .   we emphasize the difference in impact of losses in the process of state generation as opposed to those occurring in propagation channel  .   we show moreover , that correlated losses in modern realizations of schemes of large cross - kerr nonlinearity might lead to enhancement of non - classicality  . 
\\
        \midrule
      {\it Beam search}  (R-1: 47.48)& decoherence due to losses is a main obstacle to implementations of schemes using kerr and cross - kerr nonlinearities  . in our work   we discuss an aspect of the decoherence which has been seldom discussed when considering an influence of losses on states generated via kerr nonlinearity  .   namely , we address losses arising in the process of generation and not due to propagation of the generated state via lossy channels  .   we concentrate our attention on a feature that might be quite significantly pronounced in modern schemes of generating large kerr nonlinearity : the modal loss can be strongly correlated  .   indeed , the modes occupy the same volume and interact with the same physical systems which form the reservoirs  .   also , if the kerr - nonlinearity scheme implies a sufficiently strong dispersive coupling of light modes to emitters , then coupling of these emitters to dissipative reservoirs might also appear to be quite strong  . as a result , this would mean strongly correlated modal losses  . on a number of examples   we show how the correlated loss enhances and creates intermodal correlations and even entanglement , and might lead to generation of entangled states quite \dots
        \\
      \midrule
      {\it Global-aware}  (R-1: 41.50) &decoherence of quantum states generated via kerr and cross - kerr nonlinearities is discussed  .   it is shown that modal loss can be strongly correlated , i .e . , it can be decoupled from uncorrelated reservoirs  .   this strongly correlated loss enhances and creates intermodal correlations and even entanglement , and might lead to generation of entangled states quite different from those generated in the same scheme without loss  .   the strongly correlated loss can result in the significantly enhanced robustness of the generation scheme  .\\
      \bottomrule
     \end{tabular}
    
\end{table}

\begin{table}[htbp]\scriptsize
    \centering
     \caption{Generated summaries on Multi-News sampled by ROUGE-1 $F_1$ score.}
    \begin{tabular}{p{43pt}|p{330pt}}
    \toprule
        {\it Reference} &  – Joy Milne noticed a small difference in her husband just before he turned 40 . "His smell changed," the Scottish woman tells the BBC . "It wasn\'t all of a sudden . It was very subtle—a musky smell ." He was diagnosed with Parkinson\'s disease six years later and died in June at age 65 . Milne found herself supporting the Parkinson\'s UK charity, where she noticed others had the same distinct smell . She happened to mention her observation to scientists at a talk, and they later put her to the test, presenting her with T-shirts worn by six people with Parkinson\'s and six without . "She got the six Parkinson\'s, but then she was adamant one of the \'control\' subjects had Parkinson\'s," says a scientist . Researchers were impressed and dubbed her a "super-smeller," per AFP . Then eight months later, that one control subject returned to tell scientists he had just been diagnosed with Parkinson\'s . Because of Milne\'s sniffer, scientists now theorize \dots
\\
        \midrule
      {\it Beam search}  (R-1: 48.93)&– Joy Milne\'s husband, Les, was diagnosed with Parkinson\'s disease at age 45—six years before he died in June . "His smell changed and it seemed difficult to describe . It wasn\'t all of a sudden . It was very subtle—a musky smell," the Australian woman, who lives in Perth, tells the BBC . "I got an occasional smell ." After joining the charity Parkinson\'s UK, she met people with the same odor, and researchers at Edinburgh University decided to test Milne\'s ability to detect the disease using t-shirts worn by six people with Parkinson\'s and six without . "Her job was to tell us who had Parkinson\'s and who didn\'t," says Dr . Tilo Kunath . "Her accuracy was 11 out of 12 . We were quite impressed ." Six of the Parkinson\'s patients wore the shirts for a day, while the other six wore them for a week . Milne correctly identified six of the Parkinson\'s patients and six of the non- Parkinson\'s patients, but she was "adamant" that one of the non- Parkinson\'s subjects had the disease, Kunath says . Eight months later, he told Kunath he\'d been diagnosed with Parkinson\'s \dots
        \\
      \midrule
      {\it Global-aware}  (R-1: 55.18) & – When Joy Milne\'s husband was diagnosed with Parkinson\'s disease six years before his death in June, her sense of smell told her something was amiss . "His smell changed and it seemed difficult to describe . It wasn\'t all of a sudden . It was very subtle—a musky smell," she tells the BBC . After joining the charity Parkinson\'s UK, she met people with the same odor . She mentioned it to scientists at the University of Edinburgh, and they decided to test her—and were "quite impressed," says one of the scientists . Milne correctly identified six people with Parkinson\'s and six without it using T-shirts that had been worn by six people with Parkinson\'s and six without it . Her accuracy was 11 out of 12 . "She got the  \dots \\
        \toprule
         {\it Reference} & – President Obama devoted his weekly radio address today to the Connecticut school shooting, asking Americans to pray for the victims and their families . "Most of those who died were just young children with their whole lives ahead of them," he said, echoing yesterday\'s news conference . "And every parent in America has a heart heavy with hurt ." Obama reiterated that it\'s time to "come together to take meaningful action" to prevent future shootings, thought he stopped short of specifics, reports Reuters . (USA Today has the full transcript here .) Advocates of stricter gun control already have begun to step up pressure, including New York City Mayor Michael Bloomberg . "We have heard all the rhetoric before," he said yesterday . "What we have not seen is leadership—not from the White House and not from Congress . That must end today ."
\\
        \midrule
      {\it Beam search}  (R-1: 59.23)&– "Most of those who died were just young children with their whole lives ahead of them . And every parent in America has a heart heavy with hurt," President Obama said today in his weekly radio address, reports USA Today . "There are families in Connecticut who can\'t do that today . And they need all of us now . All of us can extend a hand to those in need—to remind them that we are there for them, that we are praying for them ." Obama again stopped short of calling for tougher gun laws, reports Reuters . "We grieve for the families of those we lost," he said . "And we keep in our prayers the parents of those who survived ."
        \\
      \midrule
      {\it Global-aware}  (R-1: 48.81) &– President Obama today repeated his call for "meaningful action" to reduce gun violence in the wake of the Connecticut school shooting, reports Reuters . "We grieve for the families of those we lost," he said in his weekly radio address . "And we keep in our prayers the parents of those who survived ." Obama didn\'t call for stricter gun laws, notes USA Today, but he did urge a national conversation . "This weekend, Michelle and I are doing what I know every parent is doing—holding our children as close as we can and reminding them how much we love them," he said . "There are families in Connecticut who can\'t do that today . And they need all of us now ."
\\
 \toprule
         {\it Reference} & – Jesse Jackson Jr . wept in court today, apologized for using \$750,000 in campaign funds as his personal piggy bank, and then took his lumps: 30 months in federal prison and three years\' probation, reports the Chicago Tribune . “I am the example for the whole Congress,” he said . “I understand that . I didn’t separate my personal life from my political activities, and I couldn’t have been more wrong .” The former Illinois congressman added an unusual request: He wants to serve his time in Alabama because "I want to make it a little inconvenient for everybody to get to me .” His wife, Sandi, then got sentenced to 12 months in prison for filing false tax returns about their income, reports AP . She wept, too . “My heart breaks every day with the pain this has caused my babies,” she said, referring to their two children, ages 9 and 13 . Because of the kids, the judge will \dots
\\
        \midrule
      {\it Beam search}  (R-1: 52.74)& – Former congressman Jesse Jackson Jr . and his wife, former Chicago alderman Sandi Jackson, were sentenced to prison today, reports the Chicago Sun-Times . Jesse Jackson Jr . got 2 1/2 years for using campaign funds for personal expenses, while his wife got one year for filing false tax returns . The Jacksons had pleaded for mercy for each other in court . Jesse Jackson Jr . "I didn’t separate my personal life from my political life, and I couldn’t be more wrong," he said . "I take responsibility for my actions ." Sandi Jackson said she needed to be with her children . "I ask to continue to provide for my children," she said . "To take the mother away  . . . would be an unbearable burden on these two children ."
        \\
      \midrule
      {\it Global-aware}  (R-1: 47.29) &– Former congressman Jesse Jackson Jr . and his wife, former Chicago alderman Sandi Jackson, were sentenced to prison today, reports the Chicago Sun-Times . Jackson was sentenced to 2 1/2 years for using campaign funds for personal expenses, while his wife got a year for filing false tax returns . The Jacksons had pleaded for mercy for each other in court, but the judge didn\'t seem swayed, reports AP . “There may be blurred lines for congressmen to follow when their lives are political . This case did not come near those areas,” said Judge Amy Berman Jackson . “I cannot do it and I will not do it .” She rejected Jackson Jr .\'s defense that his bipolar disorder played a role, saying his string of accomplishments—" propped up by a political family dynasty"—"points to only one conclusion, and that is that you knew better ." She also rejected \dots\\
      \bottomrule
     \end{tabular}
    
\end{table}

\clearpage
\section{ Global-aware Inference in Neural Machine Translation}\label{app:nmt}
Though the focus of this paper is Neural Abstractive Summarization, we will share how to enable global-aware inference to improve beam search significantly in NMT using a simple variation.

The gap between NMT and NAS is that NAS is a more-to-less generation task where the global attention values of many words are very low, which makes it relatively easy to trigger punishment at the beginning. As one can see in App.~\ref{app: dd} and the examples provided, many global-aware summaries are different from that of beam search from scratch, implying that the global scoring mechanism is activated in the very early stage. Since translation is a nearly one-to-one task, it is more likely to ignore some locality biases during the beginning stage when the local attention is generally low. Notably, although there is no very low global attention in translation tasks, their global attention distributions are still irrelevant with uniform distributions (here we only consider pre-trained translation models).

\begin{table}[htbp]
    \centering
     \caption{BELU results of WMT16. We perform the blocked global-aware inference with the block length of $10$.}
    \begin{tabular}{lc}
    \toprule
$\mathcal{K}=5$ & {\bf BELU}\\
    
    \midrule
    {\bf beam search} & 37.6\\
     \midrule
    {\bf ORACLE global-aware } & 39.2\\
     {\bf + Blocking} & {\bf 41.6}\\
     \midrule
     {\bf Global-aware} & 37.8\\
     {\bf + Blocking} & {\bf 38.1}\\
    
       \bottomrule
     \end{tabular}
    
    \label{tab:nmt}
\end{table}

To alleviate the gap, a straightforward-but-effective approach is to transform the one-to-one generation task to several more-to-less tasks. We divide the reference into several blocks of equal length and predict the global attention distribution of each block. Taking a reference of $25$ tokens as an example, if we set the block length as $10$, then there will be three blocks, namely the block of $1-10$ tokens, $11-20$ tokens, and the last 5 tokens. We use the special token $b_1$ to denote the first block, $b_2$ to denote the segment composed of the first and second block, and $b_0$ to denote the whole reference. When we train the attention-prediction model, these special tokens are concatenated before the source document to tell the predictor which segment it should predict. During inference, we first predict the global attention distribution of $b_0$ and calculate the optimal length, e.g. $32$, then we can figure out there are three more global attention distributions that need  be predicted, i.e., $b_1$, $b_2$ and $b_3$. We apply the predicted attention distribution of $b_1$ to guide the inference at the beginning, and the attention distribution is transformed to that of $b_2$ when the length of generated sequence is larger than $10$ and so on. When the decoding step reaches $30$, the attention distribution of $b_0$ is deployed until all candidate sequences are terminated. Although this blocking operation would increase the difficulty of training and inference, these negative effects are controllable by adjusting the block length.

We use mBART \cite{liu2020multilingual} as the neural translation model to examine the blocking operation on WMT16 en-ro sentence-level translation dataset, and related results are presented in Table \ref{tab:nmt}. Compared with beam search, our proposed global-aware  elevates the BELU score from $37.6$ to $37.8$, and ORACLE global-aware improves the result to $39.2$. The blocking operation further improves global-aware to $38.1$, and ORACLE global-aware to $41.6$. We also experiment this operation on summarization tasks where minor improvement is observed.

Finally, since the blocking operation would inevitably increase the difficulty of training and inference and is not conducive to the robustness, it may not be the most reasonable direction for improving the global-aware inference in one-to-one tasks. We think there are three following issues that might be worth exploring in the future. 1) Whether the proposed global-aware inference has already been able to improve document-level translation significantly; 2) intuitively, we averaged the global attention distributions of all layers, but it is possible that the  global attention in a single layer or a combination of several layers can better guide decoding; 3) whether a more comprehensive global protocol can be found to regulate beam search in a more detailed way.

\end{document}